\documentclass{article}

\usepackage{microtype}
\usepackage{graphicx}
\usepackage{subfigure}
\usepackage{booktabs} 
\usepackage[numbers]{natbib}
\usepackage{hyperref}
\usepackage{algorithm,algorithmic}

\usepackage[final]{neurips_2020}

\usepackage{amsmath,amsthm,amsfonts,bm}
\usepackage{graphicx}
\usepackage{subfigure}
\usepackage{algorithm,algorithmic}
\usepackage{multirow}
\usepackage{supertabular}
\usepackage{epstopdf}
\newcommand{\RR}{\mathbb{R}}

\newtheorem{theorem}{Theorem}
\newtheorem{proposition}{Proposition}
\newtheorem{remark}{\indent Remark}
\newcommand{\veps}{\varepsilon}

\DeclareMathOperator{\supp}{supp}

\DeclareMathOperator{\avg}{\mathbb{E}}
\graphicspath{{FIGS/}}

\title{Multilabel Classification by Hierarchical Partitioning\\ and Data-dependent Grouping}

\author{%
  Shashanka Ubaru\thanks{Corresponding Author. Email : \texttt{Shashanka.Ubaru@ibm.com}}\\
IBM Thomas J. Watson Research Center\\
Yorktown Heights, NY, USA\\
\And  
Sanjeeb Dash\\
IBM Thomas J. Watson Research Center\\
Yorktown Heights, NY, USA\\
   \And
   Arya Mazumdar\\
 Department of Computer Science \\
  University of Massachusetts, Amherst, MA\\
  \And
     Oktay Gunluk\\
 Operations Research and Information Engg  \\
  Cornell Universty, Ithaca, NY\\
}
\begin{document}
\maketitle
\begin{abstract}

In modern  multilabel classification problems, each data instance  belongs  to  a small number of classes from a large  set of  classes. In other words, these problems involve learning very sparse binary label  vectors. Moreover, in large-scale problems, the labels typically have certain (unknown) hierarchy.
In this paper we exploit the sparsity of label vectors and the hierarchical structure to embed them in low-dimensional space using label groupings. 
Consequently, we solve the classification problem in a much lower dimensional space and then obtain labels in the original space using an appropriately defined lifting.
Our method builds on the work of~\cite{ubaru2017multilabel}, where the idea of group testing was also explored for multilabel classification. 
We first present a novel data-dependent grouping approach, where we use a  group construction based on a low-rank Nonnegative Matrix Factorization (NMF) of the label matrix of training instances.
The construction also allows us, using recent results,  to develop a fast prediction algorithm that has a  \emph{logarithmic runtime in the number of labels}.  
We then present a hierarchical partitioning approach that exploits the label hierarchy in large-scale problems to divide up the large label space and create  smaller sub-problems, which can then be solved independently via the grouping approach.
Numerical results on many benchmark datasets illustrate that, compared to other popular methods, our proposed methods achieve competitive accuracy with significantly lower computational costs. 
\end{abstract}

\section{Introduction}
Multilabel classification (MLC) problems involve learning how to predict a (small) subset of classes a given data instance belongs to from a large set of classes.
Given a set of labeled training data $\{x_i,y_i\}_{i=1}^n$ instances with input feature vectors $x_i\in\RR^p$ and label vectors $y_i\in \{0,1\}^d$,  we wish to learn the relationship between $x_i$s and $y_i$s in order to predict the label vector  of a new data instance.
MLC problems are encountered in many domains such as recommendation systems \cite{jain2016extreme}, bioinformatics~\cite{tai2012multilabel},  computer vision~\cite{Deng2011}, natural language processing~\cite{mikolov2013distributed}, and music~\cite{trohidis2008multi}.
In the large-scale MLC problems that we are interested in, the number of labels $d$ can be as large as $O(n)$ but the $\ell_0$-norm of the label vectors is quite small (constant).
In some modern applications, the number of classes can be in the thousands, or even millions~\cite{wydmuch2018no,Jain2019slice}. However, the label vectors are typically sparse as individual  instances  belong  to just a  few  classes.  Examples of such large-scale MLC problems include  image and  video annotation for searches~\cite{wang2009multi,Deng2011},
ads recommendation  and web page categorization~\cite{agrawal2013multi,Prabhu20141fastxml}, tagging text and documents for categorization~\cite{tsoumakas2008effective,jain2016extreme}, and others~\cite{Jain2019slice}. 
There are two practical challenges associated with these large-scale MLC problems: (1) how many classifiers does one have to train, and later, (2) what is the latency to predict the label vector of a new data instance using these classifiers.  In the rest of this paper, we address these two challenges.
 
\paragraph{Related Work:}\label{sec:related} 
Most of the prior methods that have been proposed to solve large-scale sparse MLC problems fall under four categories:

(1) \emph{One versus all (OvA) classifiers}: Earlier approaches for the MLC problem involve training a binary classifier for each label independently~\cite{Zhang2018}.
Recent approaches such as DiSMEC~\cite{babbar2017dismec}, PD-Sparse~\cite{Yen2016pdsparse}, PPD-Sparse~\cite{yen2017ppd}, ProXML~\cite{babbar2019data}, and Slice~\cite{Jain2019slice} propose different paradigms to deal with the scalability issue of this naive approach. These methods typically train linear classifiers and achieve high prediction accuracy but at the same time suffer from high training and prediction runtimes. 
 Slice reduces the training cost per label by subsampling the negative training points and  reducing the number of training instances logarithmically. 

(2) \emph{Tree based classifiers}: These approaches exploit the hierarchical nature of labels when there is such a hierarchy, e.g., HOMER~\cite{tsoumakas2008effective}. 
Recent tree based  methods include FastXML~\cite{Prabhu20141fastxml}, PfastreXML~\cite{jain2016extreme}, Probabilistic Label Trees~\cite{pmlr-v48-jasinska16}, Parabel~\cite{Prabhu2018parabel},  SwiftXML~\cite{Prabhu2018swift},  extremeText~\cite{wydmuch2018no}, 
CraftXML~\cite{siblini2018craftml}, and Bonsai~\cite{khandagale2019bonsai}. 
These methods yield high prediction accuracy when labels indeed have a  hierarchical structure.
However, they also tend to have high training times as they typically use clustering methods for label partitioning, and need to train many linear classifiers, one for each label in leaf nodes. 
 
(3) \emph{Deep learning based classifiers}: More recently, neural network based methods such as XML-CNN~\cite{Liu2017XMLCNN}, DeepXML~\cite{Zhang2018Deep}, AttentionXML~\cite{you2018attentionxml}, and X-BERT~\cite{chang2019modular} have also been proposed. 
These methods  perform as well as the tree based and OvA methods in many  cases. However, they also suffer from high training and prediction costs, and the resulting model sizes can be quite large (in GBs).

(4) \emph{Embedding based classifiers}: These approaches reduce the  number of labels by projecting the label vectors onto a low-dimensional space. 
Most of these methods assume that the label matrix $Y$ is low-rank, see\cite{tai2012multilabel,bi2013efficient,zhang2011multi,chen2012feature,yu2014large}. In this case, certain  error guarantees can be established using the label correlation.
However, the low-rank assumption does not always hold, see ~\cite{bhatia2015sparse,xu2016robust,babbar2017dismec}.
Recent embedding  methods such as SLEEC~\cite{bhatia2015sparse}, XMLDS~\cite{gupta2019distributional} and DEFRAG~\cite{ijcai2019-361}
 overcome this issue by using local embeddings and negative sampling.
Most of these embedding methods require expensive techniques to recover the high-dimensional label vectors, involving eigen-decompositions or matrix inversions, and solving large optimization problems.

To deal with the scalability issue, a group testing based approach (MLGT) was recently proposed in~\cite{ubaru2017multilabel}. This method involves creating $m$ random subsets (called groups defined by a binary group testing matrix) of classes and training  $m$ independent binary classifiers to learn whether a given instance belongs to a group or not.  
When the label sparsity is $k$, this method requires only $m=O(k^2\log d)$ groups to predict the $k$ labels and therefore, only a small number of classifiers need to be trained.
Under certain assumptions, the labels of a new data instance  can be predicted by simply predicting the groups it belongs to. 
The MLGT method has been shown to yield low Hamming loss errors. 
However, since the  groups are formed in a random fashion, the individual classifiers might be poorly trained. 
That is,  the random groupings might club together unrelated classes  and the binary classifiers trained on such groups will be inefficient.

\paragraph{Our contributions:}
In this work, we build on the MLGT framework and present a new MLC approach based on hierarchical partitioning and a data-dependent group construction. We first present the novel grouping approach (NMF-GT) that improves the accuracy of MLGT. 
 This new method samples the group testing (GT) matrix (which defines the groups) from a low-rank Nonnegative Matrix Factorization (NMF) of the  training data label matrix $Y = [y_1,\ldots, y_d]$. Specifically, we exploit symmetric NMF~\cite{symmNMF}
 of the correlation matrix $YY^T$, which is known to capture the clustering/grouping within the data~\cite{dingNMF}.
  This helps us capture the label correlations in the groups formed, yielding  better trained classifiers.
 We analyze the proposed data-dependent construction and give theoretical results explaining why it performs well in MLGT. In the supplement, we discuss a GT construction that has constant weight across rows and columns, i.e., each group gets the same number of labels, and each label belongs to same number of groups. These constructions yield better classifiers and improved decoding, see Section~\ref{sec:expts} for details. 
 
 These new constructions also enable us -- using recent results -- to develop  a \emph{novel  prediction algorithm} with \emph{logarithmic} runtime in the number of labels $d$. If the sparsity of the label vector desired is $k$, then the complexity of the prediction algorithm will be  $O(k \log \frac{d}{k})$.  This significant improvement  over existing methods will allow us to predict labels of new data instances in high-throughput and real-time settings such as recommendation systems~\cite{Prabhu2018swift}. This will address some of the limitations in  traditional approaches to obtain related searches (search suggestions) ~\cite{Jain2019slice}.

  We then present a hierarchical partitioning approach that exploits the label hierarchy in large-scale problems to divide the large label set into smaller subsets. The associated sub-problems can then be solved simultaneously (in parallel) using the MLGT approach. During prediction, the outputs of individual fast decoders are simply combined (or weighted) to obtain the top $k$ labels in log time. In numerical experiments, we first show that the new group construction (NMF-GT) performs better than the previous random constructions in~\cite{ubaru2017multilabel}. 
 We then compare the performance of the proposed hierarchical method (He-NMFGT) to some of the popular  state-of-the-art methods on large datasets. We also show  how the group testing framework can achieve \emph{learning with less labeled data} for multilabel classification.

\section{MLGT method}\label{sec:MLGT}
We first describe the group testing framework for MLC
problems.
The training data consists of $n$  instances $\{(x_i,y_i)\}_{i=1}^n$, 
where $x_i\in\RR^p$ are the input feature vectors and  $y_i\in\{0,1\}^d$  are the corresponding label vectors for each instance, and are assumed to be
$k$-sparse, i.e., $||{y_i}||_0\le k$.

 \paragraph{Training.}
  The first step in training  is to  construct an $m\times d$  binary matrix $A$, called the {\em group testing} matrix. Rows of $A$ correspond to groups, columns to labels, and $A_{ij}$ is 1 if the $j$th label index (or class) belongs to the $i$th group. There exists an $A$ with $m = O(k \log d)$ (e.g., a $k$-disjunct matrix, see~\cite{ubaru2017multilabel}) such that for any $k$-sparse binary vector $\tilde{y} \in \{0,1\}^d$, $\tilde{y}$ can be uniquely recovered (in polynomial time) from $\tilde{z} = A \vee \tilde{y}$. Here $\vee$ is the Boolean OR operation (replacing the vector inner product between a row of $A$ and $\tilde{y}$ in $A\tilde{y}$).
  In section~\ref{sec:const}, we describe how to construct these group testing matrices.
  This motivates projecting the label space into a lower-dimensional space via $A$, and creating {\em reduced} label vectors  
  $z_i$ for each $y_i, i=1,\ldots,n$ where $z_i=A\vee y_i$.
  The last step is to train $m$ binary classifiers $\{w_j\}_{j=1}^m$ 
  on $\{x_i,(z_i)_j\}_{i=1}^n$ where $(z_i)_j$, the $j$th entry of $z_i$, indicates whether the $i$th instance belongs to the $j$th group or not. 
Algorithm~\ref{alg:algo1} summarizes the training algorithm.

  \begin{minipage}[t]{6.7cm}
  \vskip -0.1in
     \begin{algorithm}[H]
\caption{MLGT: Training Algorithm}
\label{alg:algo1}
\begin{algorithmic}
   \STATE {\bfseries Input:} Training data  $\{(x_i,y_i)\}_{i=1}^n$, group testing matrix
   $A\in\RR^{m\times d}$, binary classifier~$\mathcal{C}$.
 \STATE {\bfseries Output:} $m$ classifiers $\{w_j\}_{j=1}^m$.
\FOR{$i=1,\ldots, n$.}
\STATE $z_i=A\vee y_i$.
\ENDFOR
\FOR{$j=1,\ldots, m$.}
\STATE $w_j=\mathcal{C}(\{(x_i,(z_{i})_j)\}_{i=1}^n)$.
\ENDFOR
\end{algorithmic}
\end{algorithm}
  \end{minipage}
  \begin{minipage}[t]{6.7cm}
  \vskip -0.1in 
     \begin{algorithm}[H]
\caption{MLGT: Prediction Algorithm}
\label{alg:algo2}
\begin{algorithmic}
   \STATE {\bfseries Input:} Test data $x\in\RR^p$, the group testing matrix
   $A\in\RR^{m\times d}$, $m$ classifiers $\{w_j\}_{j=1}^m$, sparsity $k$.
 \STATE {\bfseries Output:} predicted label $\hat{y}$.
\FOR{$j=1,\ldots, m$.}
\STATE $\hat{z}(j) = w_j (x)$.
\ENDFOR
\STATE 
$\hat{y} = \text{fast-decode}(A, \hat{z}, k)$.
\end{algorithmic}
\end{algorithm}
\end{minipage}

\paragraph{Prediction.}
  For a new instance $x\in\RR^p$, we first use the $m$ classifiers $\{w_j\}_{j=1}^m$  to predict  a reduced label vector $\hat{z}$.
  We then apply the following simple linear decoding technique : For all $l\in[1,\ldots,d]$,
 $$ \hat y_l = \left\{
		\begin{array}{ll}
			 1 &\text{ if and only if~~}supp(A^{(l)})\subseteq supp(\hat{z})\\
			0& \text{ otherwise.}
		\end{array}
	\right.$$
Here, $supp(z):=\{i: z_i\ne 0\}$ denotes the support of the  vector $z$.
  When $A$ is $k$-disjunct~\cite{ubaru2017multilabel} and $\hat z = A \vee \hat y$ for some $k$-sparse vector $\hat y$, the above algorithm recovers $\hat{y}$.  
%
%
Unlike other embedding methods, this decoding technique does not require expensive matrix operations such as decompositions or inversion, and is linear in the number of labels $d$ using sparse matrix-vector products. 

We will next present a new construction of $A$ together with a decoding algorithm that is  logarithmic in $d$ and can be used in the last step of Algorithm~\ref{alg:algo2} in place of the linear decoder described above.
 
\section{Data dependent construction and decoding}\label{sec:const}
In ~\cite{ubaru2017multilabel}, the authors construct the group testing matrix $A$ using a uniform random construction that does not use any information about the training data. Even if two distinct classes (or label indices) are indistinguishable with respect to data instances, the columns of $A$ for these classes are different. 
We present a  novel data-dependent construction for $A$ such that "similar" classes are represented by similar columns of $A$ and show that this construction leads to much better prediction quality.
We also present a fast decoding technique. 
Consider the following metric:
\begin{equation}\label{eq:phiY}
\Phi_Y(A) = \|\frac{1}{n}YY^T-\frac{1}{m}A^TA\|_F,
\end{equation}
$YY^T$ is the label correlation matrix, also called the label co-occurence matrix~\cite{khandagale2019bonsai}. The $(i,j)$ entry of $YY^T$ is the number of training instances shared by the $i$th and $j$th classes. The entries of $A^TA$ give the number of groups shared by a pair of classes. Given a training label matrix $Y$, we construct $A$ so as to minimize $\Phi_Y(A)$, and have the groups membership structure for two similar classes be similar. See the supplement for relevant experiments. 
A completely random (disjunct) matrix is unlikely to yield low $\Phi_Y(A)$, since random grouping will not  capture the correlation between labels. However, for proper decoding, the GT matrix needs to be sparse and columns need to have low coherence.
We construct $A$ to account for both issues as follows.  

Given $Y$ and $m$ -- the number of groups -- we compute a rank $m$ symmetric Nonnegative Matrix Factorization  (symNMF) of $YY^T$ as $YY^T \approx H^TH$, where $H\in \RR^{m\times d}$ is called the basis matrix~\cite{symmNMF}. 
It has been shown that symNMF is closely related to clustering, see~\cite{symmNMF,dingNMF}. Given $Y$, the basis matrix $H$  defines the clustering within the labels.
Therefore, we use the columns of $H$ to sample $A$.

For a column $h_i$ of $H$, let $\bar{h}_i$ be the normalized column such that its entries add to 1. 
Let  $c$ be the column weights desired for $A$. For each column $i$, we  form  $\tilde{h}_i = c.\bar{h}_i$, and then  re-weight these $\tilde{h}_i$ vectors in order to avoid entries $>$ 1.
 We find all $\tilde{h}_i[j]> 1$, set these entries to $1$ and distribute the excess  sum $\sum (\tilde{h_i}[j] - 1)$ to the remaining entries. This is needed because many entries of $h_i$ will be zero.
The columns of $A$ are then sampled using the re-weighted $\tilde{h_i}$s as the sampling probability vectors. Then each column will have $c$ ones per column on average. 
We do this instead of sampling the $i$th column of $A$ as a random binary vector -- with the probability of the $j$th entry being 1 equal to $1/(k+1)$ -- as in the $k$-disjunct construction used in~\cite{ubaru2017multilabel}
.
In the supplement, we describe other constant weight constructions, where each group has the same number of labels, and each label belongs to same number of groups. 
Such constructions have been shown to perform well in the group testing problem~\cite{ubaru2016group,fast2017}. 

\begin{remark}[Choosing $c$]
In these constructions, we choose the parameter $c$ (the column sparsity or the number of ones per column) parameter using a simple procedure. For a range of $c$s we  form the matrix $A$, reduce and recover  (a random subset of) training label vectors, and choose the $c$ which yields the smallest Hamming loss error.
\end{remark}

In MLGT, for our data-dependent GT matrix, we can use the linear decoder described in section~\ref{sec:MLGT}. However, since the sampled matrix has constant weight columns, we can consider it as an adjacency  matrix of a left regular graph. 
Therefore, we can use the recent proposed SAFFRON 
construction~\cite{Saffron2016} and its fast decoding algorithm.

\paragraph{Fast decoding algorithm via. SAFFRON: }\label{sec:fast}
Recently, in~\cite{Saffron2016},  a biparitite graph based GT construction called SAFFRON (Sparse-grAph  codes Framework For gROup testiNg) was proposed, see the supplement for details. Since our NMF based construction ensures
constant weight columns, the resulting matrix $A$ can be viewed as an adjacency  matrix of a left regular graph. 
This helps us adapt the fast decoding algorithm developed for the SAFFRON construction for label prediction in our method.

We next briefly describe the decoding algorithm (an adaptation of the fast decoder presented in~\cite{Indyk2008} for sparse vector recovery). 
It has two steps, namely a \emph{bin decoder} and a \emph{peeling decoder}. The right nodes of the bipartite graph are called bins and the left nodes are called the variables.

Given the output reduced vector $z$ in the first step of prediction, the bin decoder is applied on to  $m_1$ bins $A_i$'s (these are $m_1$ partitions of $A$ as per the construction, see supplement), and all the variable nodes connected to singletons (connected to non-zero nodes) are decoded and put to a set say $D$. 
Next, in an iterative manner, a node from $D$ is considered at each iteration, and the bin decoder is applied to the bins connected to this variable node. If one of these node is a resolvable double-ton (connected to two nonzeros, but one already decoded), we can get a new nonzero variable ($y_i=1$). These decoded variables are moved from $D$ to a new set of peeled off nodes  $P$, and the newly decoded nonzero variable node, if any, is put in $D$.  
The decoder will  terminate when $D$ is  empty,  and  if the set $P$ has $k$ items, we have succeeded. For complete details, see~\cite{Saffron2016}.
The computational  complexity  of  the  decoding  scheme  is $O(k \log \frac{d}{k})$, 
see~\cite{fast2017}. Therefore, for any left-regular graph with  the SAFFRON construction and $m=O(k \log_2 d)$, the decoder recovers $k$ items in  $O(k \log d)$ time. We can use this fast decoder in the last step of Algorithm~\ref{alg:algo2} to predict the $k$ sparse label $\hat{y}$ for a given instance $x$.

\paragraph{Analysis:}
We next present an analysis that shows why the proposed data-dependent  construction will perform well in MLGT. 
Let $\tilde{H}$ be the $m \times d$ reweighted matrix derived from the label data $Y$.  $\tilde{H}$ is the potential matrix that is used to sample the $m \times d$ binary group testing matrix $A$. By construction, we know that the sum of entries in a column of $\tilde{H}$ is $c$, a constant. 

Suppose in the prediction phase, the correct label vector is $y \in \{0,1\}^d$. We know that there are at most $k$ ones in $y$, i.e., $|\supp(y)| \le k$. Then, by using the $m$ binary classifiers we obtain the reduced label vector $z$, which if the classifiers are exact, will be $z = A \vee y$. To perform the decoding for $y$ then, in effect we compute
$
b = A^Tz = A^T( A \vee y)
$
and set the top $k$ coordinates to $1$, the rest to $0$. The next result shows the effectiveness of this method.
\begin{theorem}[Sampling $A$ using $Y$]\label{theo:1}
For any $j \in \supp(y)$, $\avg[b_j] = c$, whereas, for any $j \notin \supp(y)$, $\avg[b_j] \le \sum_{i=1}^m \exp(-\langle y, \tilde{h}^{(i)}\rangle)$, where $\tilde{h}^{(i)}$ is the $i$th row of $\tilde{H}$.
\end{theorem}

The proof of this theorem is presented in the supplement. 
This result explains why our construction is a good idea. Indeed, since we generate $\tilde{H}$ in a data-dependent manner, any given label $y$ will likely have high correlations with the rows of $\tilde{H}$. As a result, the value of $b_j$ when $j$ is in the support of $y$ is much higher compared to the value of $b_j$ when $j$ is not in the support, with high probability. Therefore, choosing the top-$k$ coordinates of $b$ indeed will produce $y$. 

\section{Hierarchical approach for extreme classification}

In very large-scale MLC problems (called extreme multilabel or XML problems), the labels typically have certain
(unknown) hierarchy. By discovering and using this label hierarchy, one can 
design efficient classifiers for XML problems that have low computational cost.
A limitation of our data-dependent approach is that we perform symNMF of the correlation matrix $YY^T$. As the symNMF problem is NP-hard, and also difficult to solve for matrices with more than a few thousand columns, getting good quality classifiers for XML problems is not guaranteed. Moreover, these large matrices are unlikely to be low rank~\cite{bhatia2015sparse}. Therefore, we propose a simple hierarchical label-partitioning approach to divide the set of label classes into smaller sets, and then apply our NMF-GT method to each smaller set independently. 

Matrix reordering techniques on sparse matrices are popularly used for graph partitioning~\cite{karypis1998fast} and solving sparse linear systems~\cite{saad2003iterative}. Here, a large sparse matrix (usually the adjacency matrix of a large graph) is reordered such that the matrix/graph can be partitioned into smaller submatrices that can be handled independently. Since the label matrix $Y$ is highly sparse in XML problems and the labels have a hierarchy, the nonzero entries in $YY^T$ can be viewed as defining an adjacency matrix of a sparse graph. Let $G=(V,E)$ denote a graph, where each node corresponds to a label, and $e=ij \in E$ if and only if $YY^T_{ij} \neq 0$. In other words, an edge between nodes/labels $i$ and $j$ is present if and only if labels $i$ and $j$ occur together in at least one data point, which indicates ``interaction" between these labels.

Suppose that $G$ has say $\ell$ components, i.e., it can be partitioned into $\ell$ disjoint sub-graphs, as assumed in Bonsai~\cite{khandagale2019bonsai}. Then each component corresponds to a subset of labels that interact with one another but not with labels in other components. Permuting the labels so that labels in a component are adjacent to one another, and applying the same permutation to the columns of $Y$, one can obtain a block-diagonal reordering of the label matrix $YY^T$. Now the symNMF problem for $YY^T$ can be reduced to a number of smaller symNMF problems, one for each block of the matrix.
Most large datasets (label matrices) with hierarchy will have many smaller non-interacting subsets of labels and few subsets that interact with many other labels. A natural approach is to use the vertex separator partitioning  based reordering~\cite{gupta1997fast} or nested dissection~\cite{karypis1998fast} to obtain this permutation.

The idea is to find a small vertex separator $S$ of $G$ (here $S \subset V$) such that $G \setminus S$ has a number of disjoint components $C_1, \ldots, C_{\ell}$. The labels can then be viewed as belonging to one of the subsets $S \cup C_1, \ldots, S\cup C_{\ell}$, and we can apply NMF-GT to each separately. This idea can be further extended to a hierarchical partitioning of $G$ (by finding partitions of the subgraphs $C_i$ as $S_i, C_{i1},\ldots,C_{i\ell}$ -- where $S_i$ is a vertex separator of $C_i$).
Each level of the hierarchy would be partitioned further till the components are small enough so that the MLGT (sym-NMF) algorithm can be efficiently applied.

\begin{figure*}[tb!]
\centering
\includegraphics[height=0.165\textwidth,trim={2cm 2cm 2cm 2cm}]{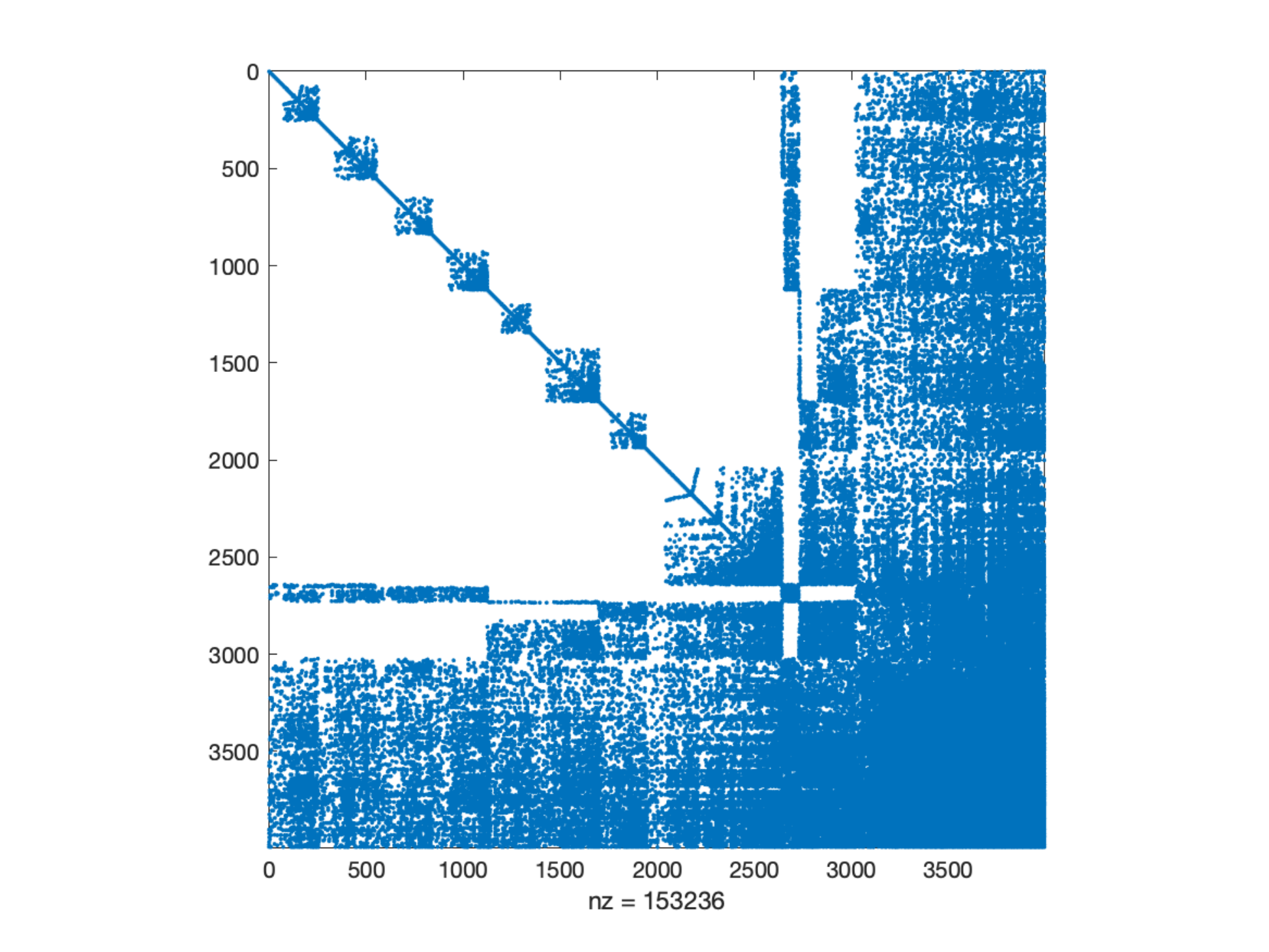}
\includegraphics[height=0.17\textwidth,trim={2cm 2cm 2cm 2cm}]{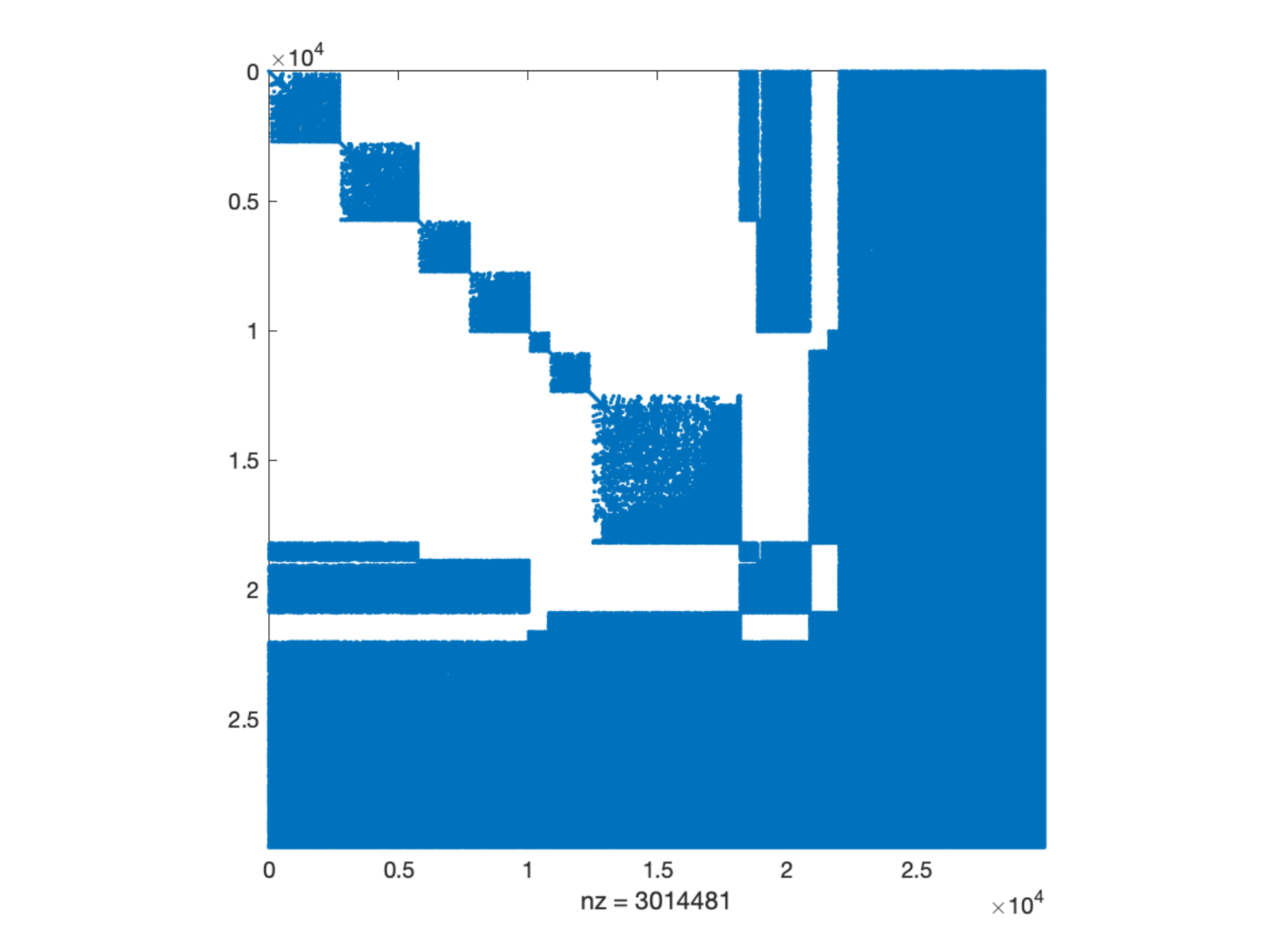}
\includegraphics[height=0.17\textwidth,trim={2cm 2cm 2cm 2cm}]{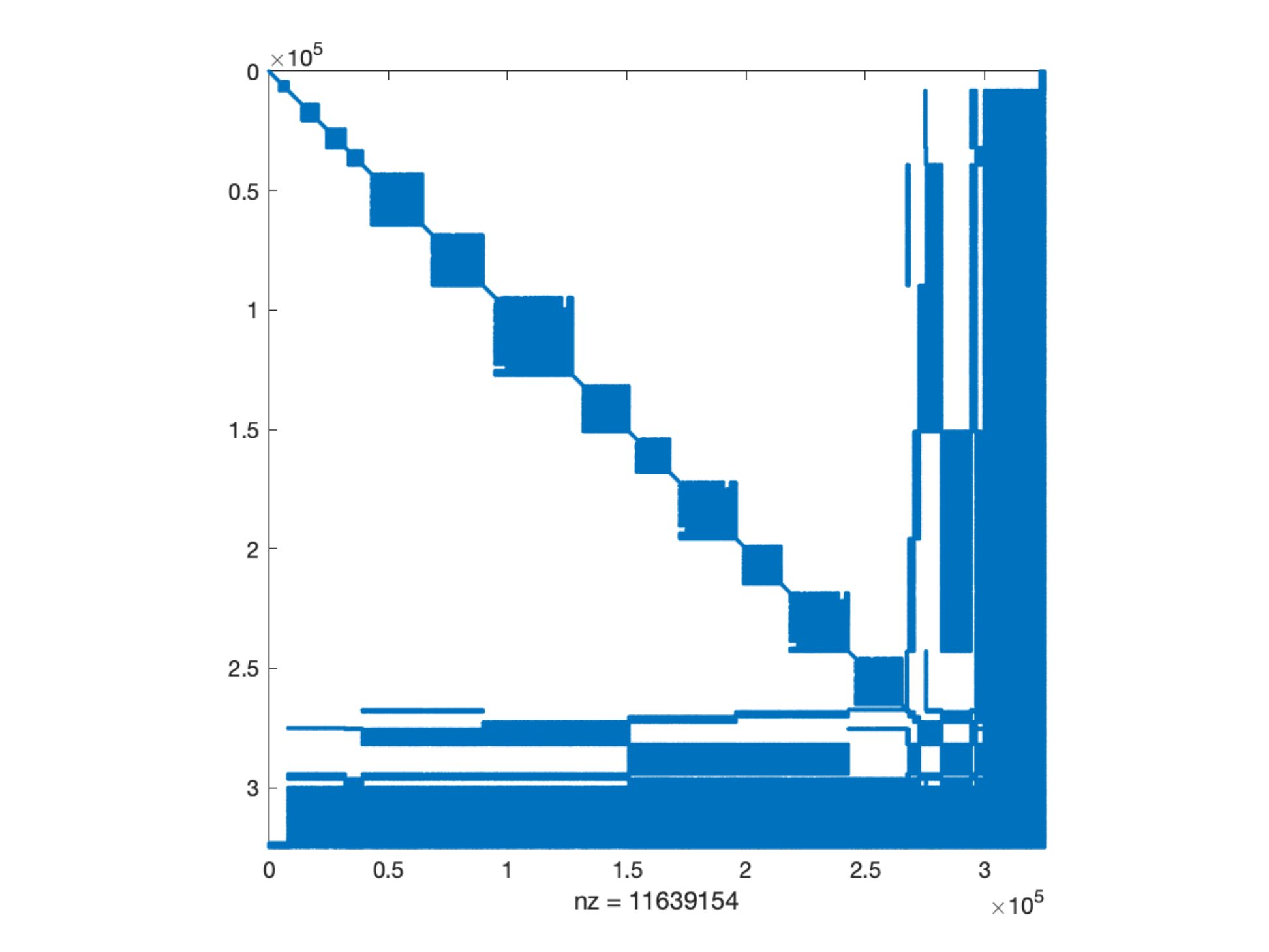}
\includegraphics[height=0.165\textwidth,trim={2cm 2cm 2cm 2cm}]{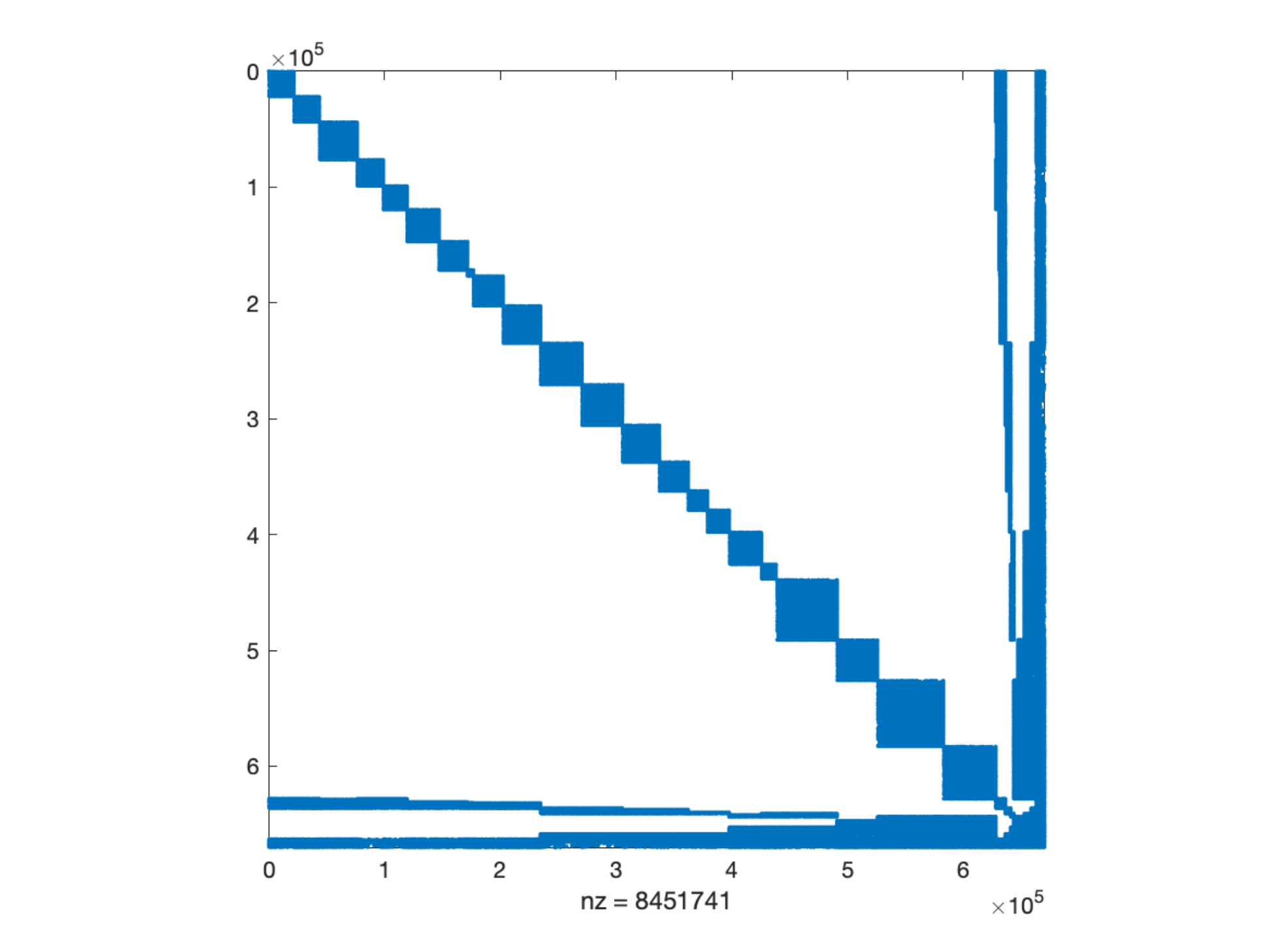}
\caption{Hierarchical reordering of label co-occurrence matrices for four XML datasets.} 
\label{fig:spy}
\vskip -0.1in
\end{figure*}

In Figure~\ref{fig:spy}, we display the hierarchical reordering of $YY^T$ obtained by the  algorithm in~\cite{gupta1997fast} for four popular XML datasets: 
Eurlex (with number of labels $d\approx 4K$), Wiki10 ($d \approx 30K$), WikiLSHTC ($d \approx 325K$), and Amazon ($d \approx 670K$), respectively. We note that there are a few distinct blocks $C_i$ (the block diagonals), where the labels only occur together and are independent of other blocks (do not interact). We also have a small subset of labels $S$ (the outer band) that interact with most blocks $C_i$. We can partition the label set into $\ell$ subsets $\{S \cup C_i\}_{i=1}^{\ell}$ of size $\{d_i\}_{i=1}^{\ell}$ each and apply our NMF based MLGT individually (it can be done in parallel). During prediction, the individual fast decoders will return the positive labels for each subsets in $O(\log d_i)$ time. We can simply combine these positive labels or weight them to output top $k$ labels. 
Since the subset $S$ of labels interact with most other labels and occur more frequently (power-law distribution), we can rank them higher when picking top $k$ of the outputted positive labels.

{\bf Comparison with tree methods:}  The tree based methods such as HOMER~\cite{tsoumakas2008effective}, Parabel~\cite{Prabhu2018parabel}, Bonsai~\cite{khandagale2019bonsai}, and others use label partitioning to recursively construct label tree/s with  pre-specified number of labels in leaf nodes or at a tree depth. Most methods use k-means clustering for partitioning (typically clustering is used to partition at each non-leaf node), and this can be expensive for large $d$.
Then OvA classifiers are learned for each label in the leaf nodes.
However, in our approach, we use label partitioning to identify label subsets on which we can apply NMF-GT independently. 
 Our matrix reordering approach is inexpensive with cost $O(nnz(YY^T)) = O(dk)$, see~\cite{gupta1997fast}. We use the NMF-GT strategy to learn only $O(k\log d_i)$ classifiers per partition.
  
\section{Numerical Experiments}\label{sec:expts}

We now present  numerical results to illustrate the performance of the proposed  approaches (the data-dependent construction NMF-GT and with hierarchical partitioning He-NMFGT)
on  MLC problems. 
Several additional results and details are presented in the supplement.

\begin{table}[H]
 \caption{Dataset statistics}\label{tab:data}
 {\footnotesize
 \begin{center} 
 \begin{tabular}{l|c|c|c|c|c}
\hline
Dataset&$d$& $\bar{k}$&$n$&$nt$ & $p$\\
\hline
Mediamill 	&101&4.38&30993&12914&120\\
Bibtex 		&159&2.40&4880&2515&1839\\
RCV1-2K (ss)	&2016&4.76&30000&10000&29699\\
EurLex-4K 	&3993&5.31&15539&3809&5000\\
AmazonCat(ss)&7065&5.08&100000&50000&57645\\
Wiki10-31K 	&30938&18.64& 14146& 6616&101850\\
WikiLSHTC  &325056&3.18& 1813391& 78743&85600\\
Amazon-670K&670091&5.45&490449&153025&135909\\
\hline
\end{tabular}
 \end{center} 
 }
\end{table}

\paragraph{ Datasets:} 
For our experiments, we consider some of the popular publicly available multilabel datasets 
put together in \emph{The Extreme Classification
Repository}~\cite{bhatia2015sparse} (\url{http://manikvarma.org/downloads/XC/XMLRepository.html}).
The applications, details and  the original sources of the datasets can be found
 in the repository. Table~\ref{tab:data} lists the statistics.

 In the table, $d=\#$labels, $\bar{k}=$average sparsity per instance, $n=\#$ training instances, $nt=\#$ test instances and 
 $p=\#$features. 
The datasets marked (ss) are subsampled version of the original data with statistics as indicated. 

\paragraph{Evaluation metrics:} To compare the performance of the different MLC methods,
we use the most popular evaluation metric called  $Precison@k$ (P@k)~\cite{agrawal2013multi} 
with $k=\{1,3,5\}$.   It has been argued that this metric is more suitable for modern applications such as tagging or recommendation, where  one is interested in only predicting a subset of (top $k$) labels   correctly. 
P@k is defined as:
  \[
  P@k :=~~~\frac{1}{k} \sum_{l\in rank_k (\hat{y})} y_l,
\]
where $\hat{y}$ is  the predicted vector and  $y$ is the actual label vector.
This metric assumes that the  vector $\hat{y}$ is real valued and its coordinates can be ranked so that the summation above can be taken over the highest ranked $k$ entries of $\hat y$. For the hierarchical approach, we weight and rank the labels based on repeated occurrence (in the overlapping set $S$).

In general, MLGT method returns a binary label vector $\hat{y}$ of predefined sparsity,  there is no ranking among its non-zero entries. Hence, we also use a slightly modified  definition:
  \begin{equation}\label{eq:Pk}
  \Pi@k := \frac{1}{k} \min\big(k, \sum_{l\in top_5 (\hat{y})} y_l\big),
\end{equation}
where $top_5(\hat{y})$ is the $5$ nonzero co-ordinates of  $\hat{y}$ predicted by MLGT assuming that the predefined sparsity is set to 5. To make the comparison fair for other (ranking based) methods, we sum over the top 5 labels based on their ranking  (i.e. we use $rank_5$ instead of $rank_k$ in the original definition). 

\begin{figure*}[tb!]
\centering
\includegraphics[height=0.3\textwidth]{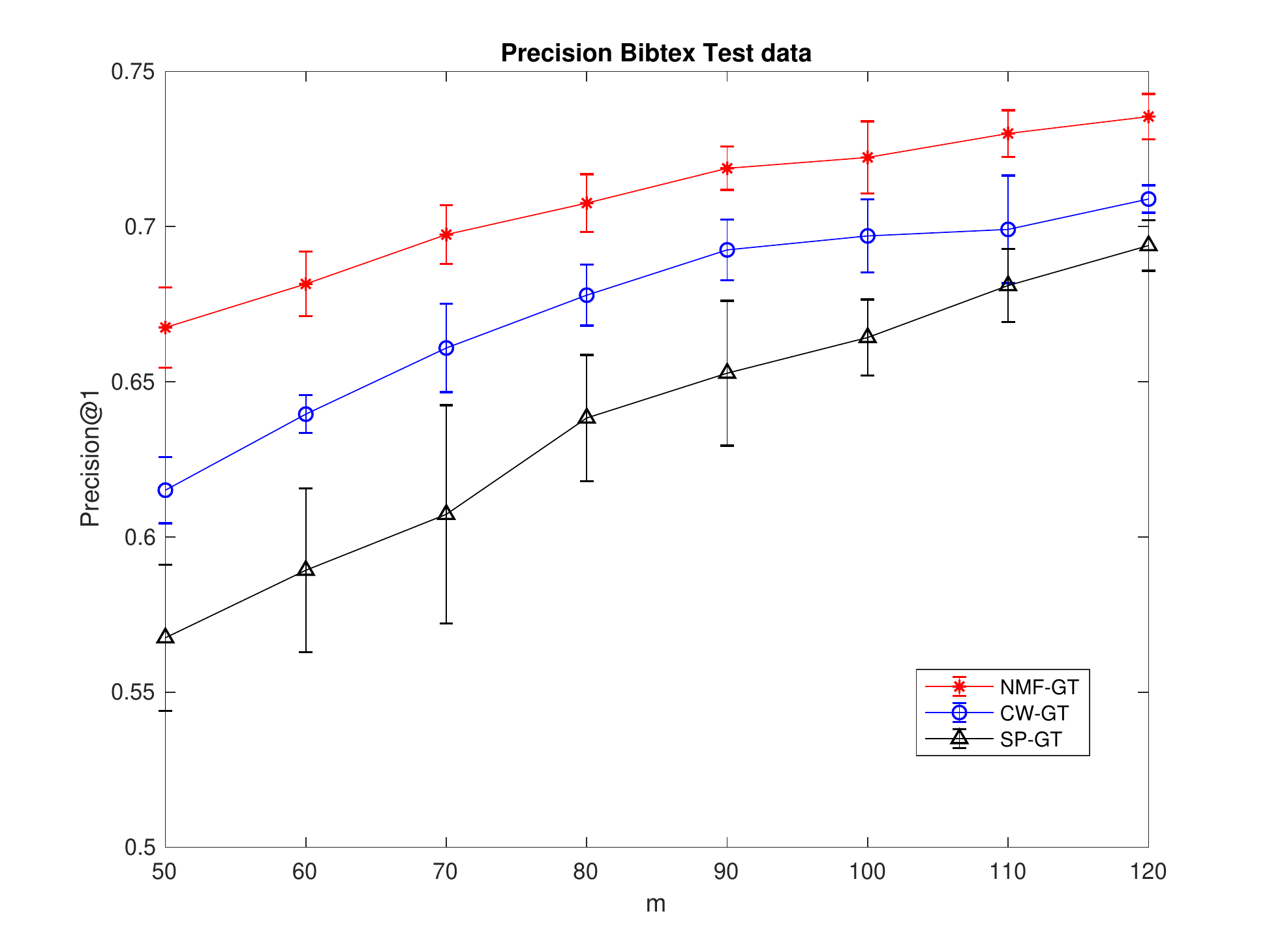}
\includegraphics[height=0.3\textwidth]{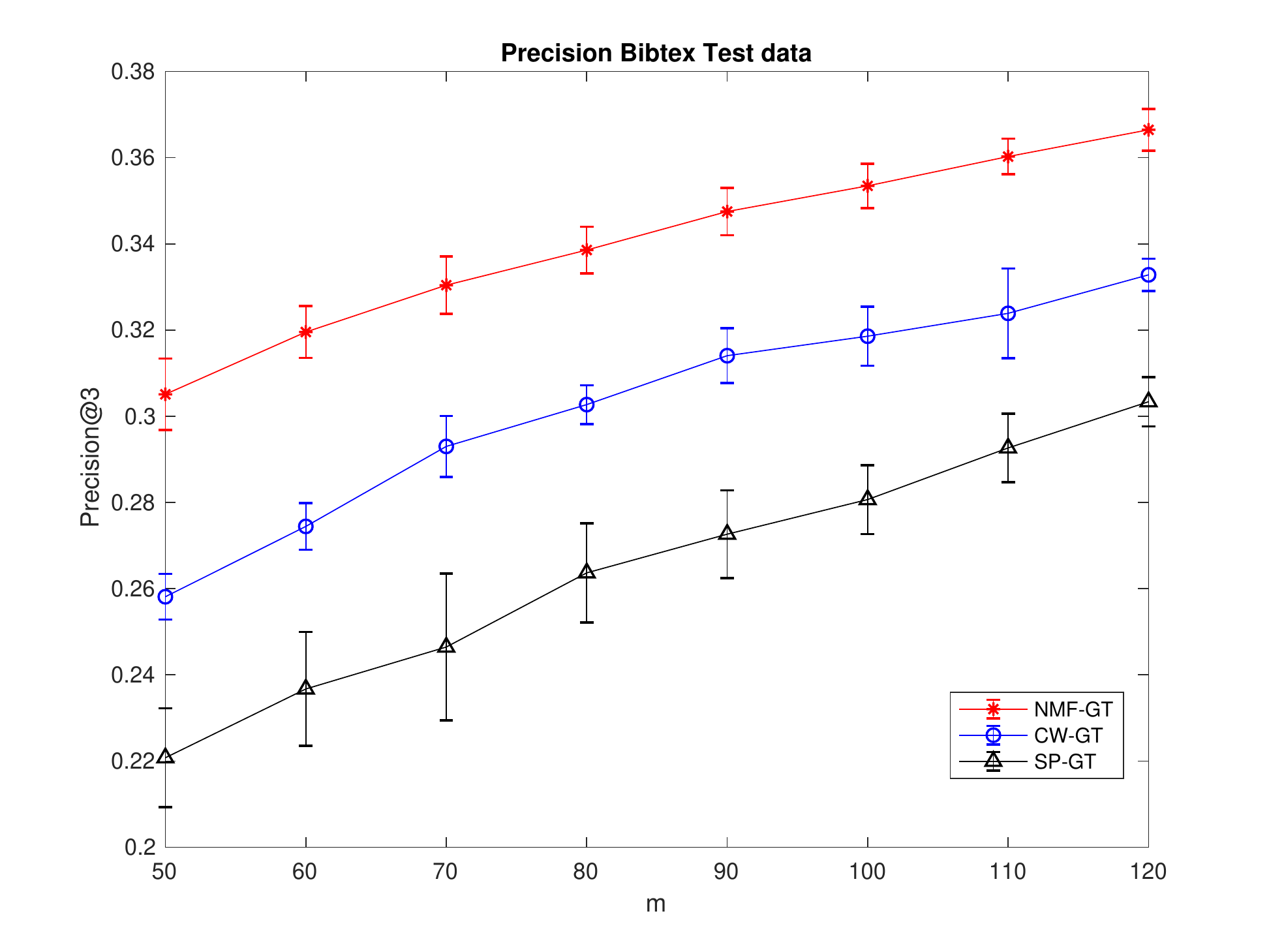}

\includegraphics[height=0.3\textwidth]{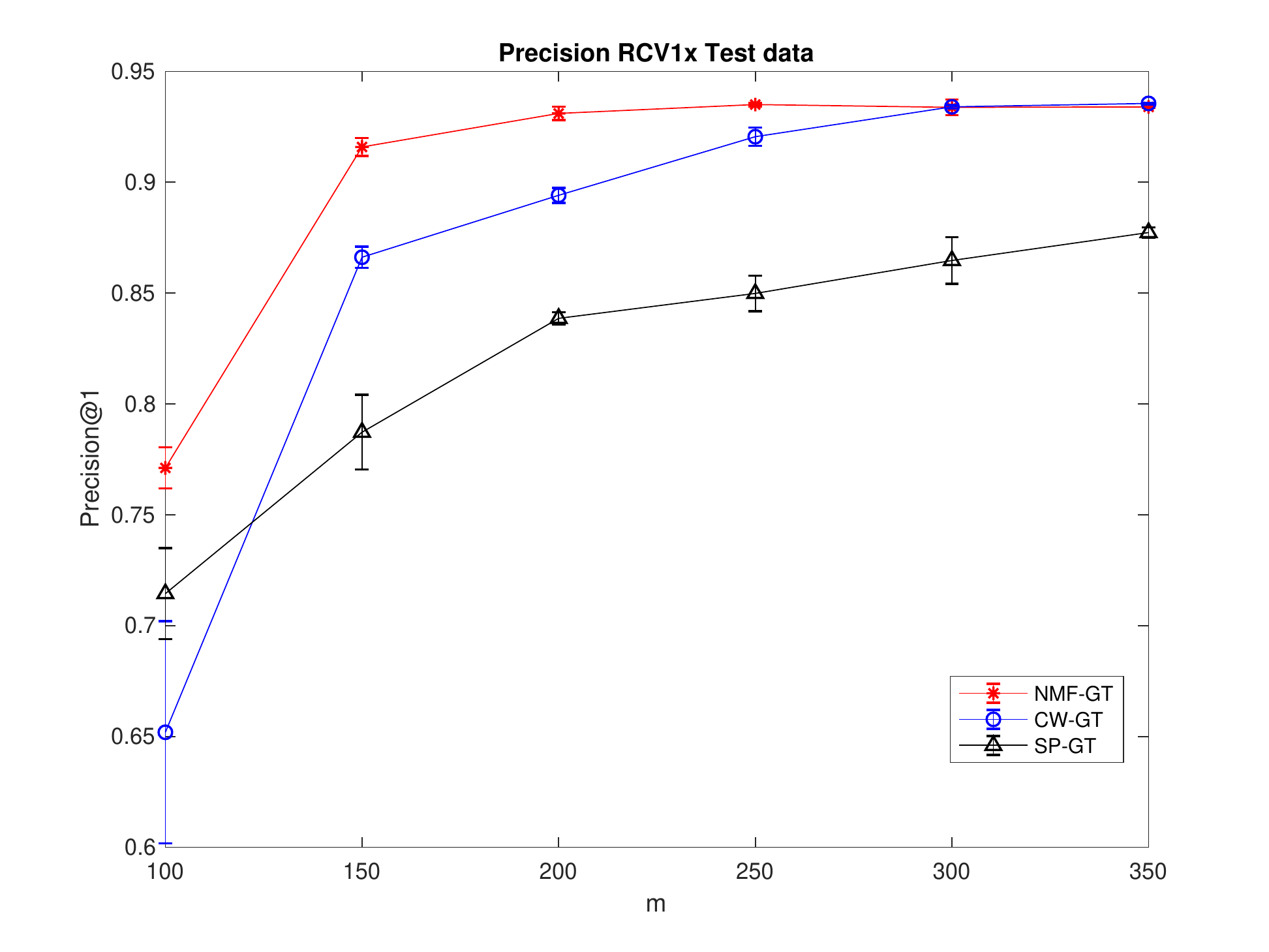}
\includegraphics[height=0.3\textwidth]{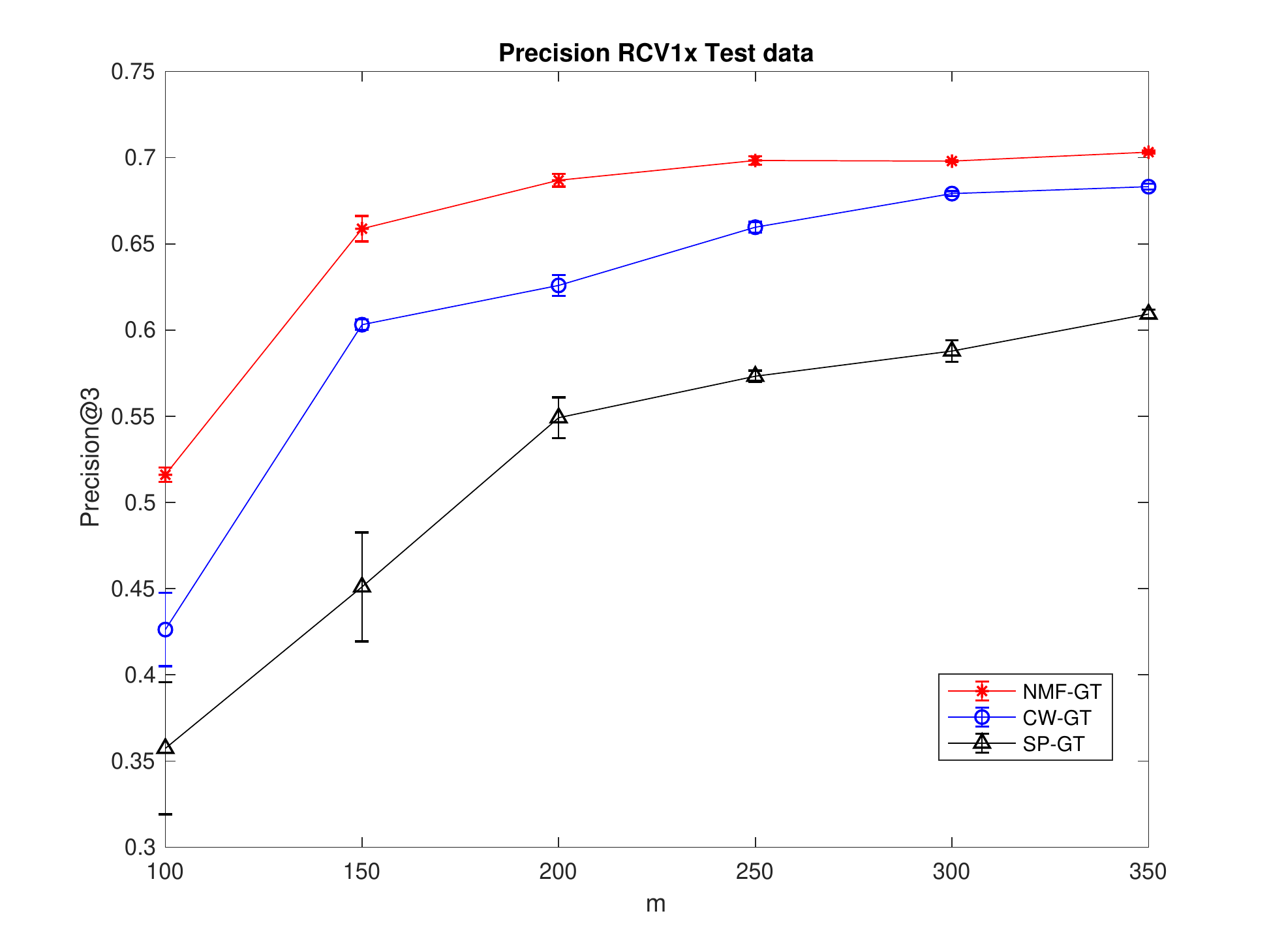}
\caption{$\Pi@1$ and $\Pi@3$ for  test data instances for bibtex (top two) and RCV1x (bottom two) datasets as a function of number of groups $m$. Error bar over 10 trials.} 
\label{fig:1}
\vskip -0.1in
\end{figure*}

\begin{table*}[tb!]
\caption{Comparisons between GT constructions. Metric: Modified Precision}
\label{tab:table1}
 \begin{center}  
\resizebox{0.6\textwidth}{!}{
\begin{tabular}{|c|c|c c c ||c|}
 \hline
  \hline
 Dataset		&Metrics & 	NMF - GT & CW - GT & SP - GT & OvA\\
 \hline \hline
 Bibtex  	   	& $\Pi@1$ &     \bf 0.7354	& 0.7089	& 0.6939	& 0.6111   \\
$d = 159$		& $\Pi@3$ &   \bf 0.3664	& 0.3328	& 0.3034  & 0.2842	\\
 $m = 120$	& $\Pi@5$ & 	\bf 0.2231	& 0.2017 	& 0.1823  & 0.1739	\\
			&$\Phi_Y(A)$&  \bf 10.610	& 12.390	& 12.983  & ---		\\
			& $T_{total}$ & 	5.13s	& 4.01s    & \bf 3.98s    & 8.22s	\\
			& $T_{test}$& \bf 0.13s		& \bf0.13s	& \bf0.13s	& 0.18s	\\
			
\hline
Mediamill  	& $\Pi@1$ &  \bf	0.8804	& 0.8286	& 0.6358	& 0.8539  \\
 $d = 101$	& $\Pi@3$ &  \bf	0.6069	& 0.5413	& 0.2729 & 0.5315	\\
$m = 50$		& $\Pi@5$ & 	\bf	 0.3693	&0.3276 	& 0.1638 & 0.3231	\\
			&$\Phi_Y(A)$&\bf 10.377	& 11.003	& 10.876  & ---		\\
			& $T_{total}$ & 17.2s	&\bf 15.7s    & 15.82s    & 29.4s	\\
			& $T_{test}$&  \bf0.17s	& \bf0.17s	& \bf0.17s	& 0.54s	\\
			
\hline
RCV1x	 	& $\Pi@1$ &  \bf	0.9350	& 0.9205	& 0.8498& 0.9289  \\
$d = 2016$ 	& $\Pi@3$ &  \bf	0.6983	& 0.6596 	&0.5732 & 0.6682	\\
$m = 250$ 	& $\Pi@5$ & 		0.4502	& 0.4104	&0.3449 &\bf 0.4708 \\
			&$\Phi_Y(A)$&\bf 53.916	& 58.459	& 58.671  & ---		\\
			& $T_{total}$ & 88.4s	& 77.5s    &\bf 74.2s    & 363.2s	\\
			& $T_{test}$&  1.20s		&\bf 1.04s	& 1.10s	& 6.37s	\\
			
\hline
Eurlex	 	& $\Pi@1$ &  	0.8477	& 0.8430	& 0.6792& \bf0.8535  \\
$d = 3993$ 	& $\Pi@3$ &  	0.5547	& 0.5582 	&0.3933 &\bf 0.6132	\\
 $m = 350$	& $\Pi@5$ & 		0.3444	& 0.3597	&0.2758 &\bf 0.4085 \\
			&$\Phi_Y(A)$&80.023	& 80.732	& 82.257  & ---		\\
			& $T_{total}$ & 227.3s	& 99.6s    & \bf90.4s    & 560.1s	\\
			& $T_{test}$& \bf 0.94s		& \bf0.93s	& \bf0.93s	& 7.26s	\\
			
\hline
\end{tabular}
}
\end{center}
\end{table*}

\paragraph{ Comparing  group testing constructions:}
In the first set of experiments, we compare the new group testing constructions with the sparse random construction (SP-GT) used in~\cite{ubaru2017multilabel}, where each entry of $A$ is sampled with uniform probability $prob = \frac{1}{k+1}$. 
Our first construction (NMF-GT) is based on the symNMF as described in Section~\ref{sec:const}.
Given the training label matrix $Y$, we first compute  the symNMF of $YY^T$ of rank $m$ using the Coordinate Descent algorithm by~\cite{symcord} (code provided by the authors) and then  compute a sparse binary matrix  using reweighted rows of the NMF basis.
Our second construction (CW-GT) is the constant weight construction defined in supplementary~\ref{sec:random}. For both constructions, the number of nonzeros (sparsity) per column of $A$ is selected using the search method described in Remark 1, see supplement for more details.

Figure~\ref{fig:1} plots  $\Pi@1$ and $\Pi@3$ we obtained for the three constructions (red star is NMF-GT, blue circle is CW-GT, and black triangle is SP-GT) as the number of groups $m$ increases. 
The first two plots correspond to the Bibtex dataset, and the next two correspond to RCV1x dataset. 
As expected,  the performance of all constructions improve as the number of groups increase. Note that NMF-GT consistently outperforms the other two.  In the supplement, we compare the   three constructions (for accuracy and runtime) on  four  datasets. We also include the One versus All (OvA) method (which is computationally expensive) to provide a frame of reference.

In Table~\ref{tab:table1}, we compare the three constructions discussed in this paper on  four  datasets. We also include the One versus All (OvA) method (which is computationally very expensive) to provide a frame of reference. In the table, we list $P@k$ for $k=\{1,3,5\}$, the correlation metric $\Phi_Y(A)$, the total time $T_{total}$ as well as the time $T_{test}$ taken to predict the labels of $nt$ test instances. 
The NMF-GT method performs better than both methods, because it groups the labels based on the correlation between them. 
This observation is supported by the fact that the correlation metric $\Phi_Y(A)$ of NMF-GT is the lowest among the three methods. 
Also note that even though NMF-GT has longer training time compared to the other GT methods (due to the NMF computation), its prediction time is essentially the same.
We also note that the runtimes of all three MLGT methods are much lower than OvA, particularly for  larger datasets as they require much fewer ($O(\log d)$) classifiers. 

In all cases, NMF-GT outperforms the other two (possibly because it groups the labels based on the correlation between them), and CW-GT performs better than SP-GT. Both NMF-GT and CW-GT ensure that $m$ classifiers are trained on similar amounts of data. 
Decoding will also be efficient since all columns of $A$ have the same support size. NMF-GT is superior to the other two constructions, and therefore, we will use it in the following experiments for comparison with other popular XML methods.

\begin{table*}[tb!]
\caption{Comparisons between different MLC methods. Metric: Modified Precision}
\label{tab:table2}
\vskip -0.1in
 \begin{center}  
\resizebox{0.8\textwidth}{!}{
\begin{tabular}{|c|c| c c c c c c|}
 \hline
  \hline
 Dataset		&Metrics & He-NMFGT &	NMF-GT & MLCS & SLEEC* & PfastreXML & Parabel\\
 \hline \hline
Mediamill  	& $\Pi@1$ & -- 	&0.8804	& 0.8359	& 0.8538& \bf0.9376 & 0.9358 \\
$d=101$		& $\Pi@3$ & -- 	&0.6069	& 0.6593	& 0.6967 &\bf 0.7701	& 0.7622\\
 $m = 50$	& $\Pi@5$ & --	& 0.3693&0.4102	&  \bf 0.5562&    0.5328	& 0.5169\\
& $T_{total}$ & -- &\bf 17.2s	& 20.3s    &   3.5m& 190.1s	& 74.19s\\
& $T_{test}$& -- &\bf  0.17s	& 6.93s	&80.5s	& 18.4s	& 17.85s\\	
\hline

RCV1x	 	& $\Pi@1$ & --  &0.9350	& 0.9244	& 0.9034	&   0.9508  &\bf  0.9680\\
$d = 2016$	& $\Pi@3$ & --   &	0.6983	& 0.6945	& 0.6395	 & 0.7412	&\bf 0.7510\\
 $m = 250$	& $\Pi@5$ & --	&	0.4502	& 0.4486	&  0.4457	& 0.4993	& \bf0.5040\\
& $T_{total}$ & -- &\bf 88.4s	& 541.1s    &  34m   &7.73m	& 6.7m	\\
& $T_{test}$& -- &\bf 1.04s	& 176.7s	& 53.1s	& 3.03m	& 1.68m	\\
		
\hline
Eurlex	 	& $\Pi@1$ &  \bf 0.9265		&0.8477	& 0.8034	& 0.7474	& 0.9004  & 0.9161\\
$d=3993$	& $\Pi@3$ &0.7084	&0.5547	& 0.5822 	& 0.5885	& 0.6946	&\bf0.7397\\
$m = 350$ 		& $\Pi@5$ &0.4807	&0.3444	& 0.3965	& 0.4776	& 0.4939 	& \bf0.5048\\
$\ell=4$			& $T_{total}$ &322s&\bf 227.3s	& 343.3s    &  21m  & 11.8m	& 6.1m\\
			& $T_{test}$ &1.1s &\bf  0.94s	& 235.1s	&  45s	& 59.2s	& 74.3s\\
			
\hline
Amazon13	 & $\Pi@1$ &\bf 0.9478 &	0.8629	& 0.7837	& 0.8053	& 0.9098  &  0.9221 \\
$d=7065$		& $\Pi@3$ & 0.6555 &	0.5922	& 0.5469 	& 0.5622	& 0.6722 & \bf 0.6957\\
 $m=700$	& $\Pi@5$ & 0.4474	&	0.3915	& 0.3257	& 0.4152	&  0.5119	& \bf 0.5226\\
$\ell = 4$			& $T_{total}$ & 8.7m &\bf 7.5m	& 19.7m    &  68.8m  & 27.5m	& 16.9m\\
			& $T_{test}$& 4.42s&\bf  4.21s	& 13.7m	& 106.3s 	& 241.6s	& 114.7s\\
			
\hline
Wiki10	 	 & $\Pi@1$ 	& \bf 0.9666  &	0.9155	& 0.5223	& 0.8079	& 0.9289  & 0.9410 \\
$d=30938$	& $\Pi@3$ 	& \bf0.7987 &0.6353	&0.2995	& 0.5050	 & 0.7269   &  0.7880\\
 $m=800$	& $\Pi@5$ 	& \bf0.5614 &0.4105	& 0.1724	& 0.3526	& 0.5061 	&0.5502 \\
$\ell=5$			& $T_{total}$ &14.7m&\bf 13.6m&  63m	 & 54.9m   & 40.5m	& 33.5m\\
			& $T_{test}$ &11.5s&\bf  9.82s& 45m	& 51.3s	& 8.2m 	& 4.2m\\
			
\hline

\end{tabular}
}
\end{center}
\vskip -0.1in
\end{table*}

\begin{table*}[tb!]
\caption{Comparisons between different XML methods. Metric: Standard Precision}
\label{tab:table3}
\vskip -0.2in
 \begin{center}  
\resizebox{\textwidth}{!}{
\begin{tabular}{|c|c| c c |c c c | c c| c|}
 \hline
  \hline
&  &  \multicolumn{2}{c|}{Embedding} & \multicolumn{3}{c|}{Tree} & \multicolumn{2}{c|}{OvA} & DNN\\\cline{3-10}
 Dataset		&Metrics & He-NMFGT  & SLEEC & PfastreXML & Parabel &XT & Dismec  & PPD-sparse & XML-CNN\\
 \hline \hline
Eurlex	 	&$P@1$ (\%)& 75.04 & 74.74 & 73.63 & 74.54	& --&   83.67&83.83 & 76.38\\
$d=3993$	& $P@3$ (\%)& 61.08 & 58.88& 60.31 & 61.72	& --&  70.70&70.72& 62.81\\
$\ell =4$	& $P@5$  (\%)& 48.07	& 47.76	& 49.39	& 50.48	& --& 59.14& 59.21& 51.41\\
$m=300$&$T_{train}$ &\bf4.8m&20m&10.8m& 5.4m & -- &0.94hr& 0.15hr& 0.28hr\\
  & $T_{test}/nt$ &\bf 0.28ms&4.87ms&1.82ms&0.91ms& --&7.05ms &1.14ms & 0.38ms\\
			
\hline
Wiki10	 & $P@1$ (\%) & 82.28 & 80.78 & 82.03 & 83.77 & 85.23 & 85.20& 73.80 & 82.78\\
$d=30938$& $P@3$ (\%)& 69.68  & 50.50 & 67.43 & 71.96 & 73.18 & 74.60& 60.90 & 66.34\\
$\ell = 5$ 	 & $P@5$  (\%) & 56.14  & 35.36 & 52.61 & 55.02 & 63.39 &  65.90& 50.40 & 56.23\\
$m=650$	& $T_{train}$ & \bf14.2m & 53m &32.3m & 29.3m& 18m&  --   & -- & 88m\\
  & $T_{test}/nt$ &\bf0.69ms &7.7ms & 74.1ms& 38.1ms& 1.83ms& --   & -- & 1.39s \\
			
\hline
WikiLSHTC	& $P@1$ (\%) &55.62  & 54.83 & 	56.05& 	64.38& 58.73&  64.94& 64.08& --\\
$d=325056$	&  $P@3$ (\%) & 33.81 & 33.42 &  36.79& 	42.40& 39.24&  42.71& 41.26& --\\
$\ell = 12$ 	    &  $P@5$  (\%) & 23.04 & 23.85 & 	27.09&	31.14& 29.26&  31.5& 30.12& --\\
$m = 800$	    & $T_{train}$ & \bf 47.5m & 18.3hr    &  7.4hr&  3.62hr & 9.2hr&  750hr& 3.9hr& --\\
      & $T_{test}/nt$ & \bf 0.8ms &   5.7ms & 	2.2ms&  1.2ms &  \bf 0.8ms&  43m& 37ms&--\\
			
\hline
Amazon670K & $P@1$ (\%)& 39.60& 35.05 & 39.46 & 43.90 & 39.90 & 45.37 &45.32& 35.39 \\
$d=670091$ &  $P@3$  (\%)&36.78 & 31.25 & 35.81&39.42 & 35.60  & 40.40 &40.37& 33.74 \\
$\ell = 22$ &  $P@5$  (\%)& 32.40& 28.56 & 33.05 & 36.09 & 32.04  & 36.96 &36.92&  32.64\\
$m = 800$	& $T_{train}$ & \bf47.8m &11.3hr & 1.23hr & 1.54hr& 4.0hr &  373hr &1.71hr& 52.2hr\\
  & $T_{test}/nt$ &\bf 1.45ms & 18.5ms & 19.3ms & 2.8ms & 1.7ms  & 429ms& 429ms& 16.2ms\\
			
\hline
\end{tabular}
}
\end{center}
\vskip -0.1in
\end{table*}

\paragraph{Comparison with popular methods:}
We next compare the NMF-GT method (best one from our previous experiments)
and the hierarchical method (He-NMFGT) with four popular methods, namely MLCS~\cite{hsu2009multi}, SLEEC~\cite{bhatia2015sparse},   PfastreXML~\cite{jain2016extreme}, and  Parabel~\cite{Prabhu2018parabel} with respect to the modified precision $\Pi@k$ metric. 
Table~\ref{tab:table2} summarizes the results obtained by these six methods for different datasets along with total computation time $T_{total}$ and the test prediction time $T_{test}$.
The no. of groups $m$ used in NMFGT and no. of blocks $\ell$ used in He-NMFGT are also given. 

We note that NMF-GT performs fairly  well given its low computational burden. The hierarchical approach He-NMFGT yields superior accuracies  with similar runtimes as NMFGT (outperforms other methods wrt.  $\Pi@1$). 
PfastreXML and Parabel yield slightly more accurate results in some cases, but require significantly longer run times.   Note that the prediction time $T_{test}$ for our methods are orders of magnitude lower in some cases. For He-NMFGT, $T_{total}$  includes computing the partition, applying MLGT for one block (since this can be done in parallel), and predicting the labels of all test instances. For smaller two datasets, He-NMFGT was not used since they lacked well-defined partitions. In He-NMFGT, for the labels shared across partitions, we  use weights for the label outputs such that these weights add to 1. The matrix reordering method recursively partitions the labels, hence discovering a hierarchy. The code we use produces the partitions (and sizes), in addition to the permutations depicted in Fig.~1.

In Table~\ref{tab:table3} we compare the performance of He-NMFGT with several other popular XML methods wrt. the standard $P@k$ metric. 
We compare the accuracies and computational costs for He-NMFGT, SLEEC (embedding method), three tree methods (PfastreXML, Parabel, ExtremeText XT), two OvA methods (Dismec, PD-sparse) and a DNN method XML-CNN (see sec.~\ref{sec:related} for references). The precision results and the runtimes for the four additional methods  were obtained from~\cite{Prabhu2018parabel, wydmuch2018no}. In the table a `--' indicate these results were not reported by the authors.

We note that, compared to other methods, He-NMFGT is significantly faster in both training and test times, and yet yields comparable results. 
The other methods have several parameters that need  to be  tuned. 
More importantly,  the main routines of most other methods are written in C/C++, while  He-NMFGT was implemented in Matlab and hence we believe the run times can be improved to enable truly real-time predictions.
The code for our method is publicly available at \url{https://github.com/Shashankaubaru/He-NMFGT}.
Several additional results, implementation details and result discussions are given in the supplement.
 
 \begin{figure}
{\centering
\includegraphics[height=0.23\textwidth,trim={1cm 7cm 1cm 7cm}]{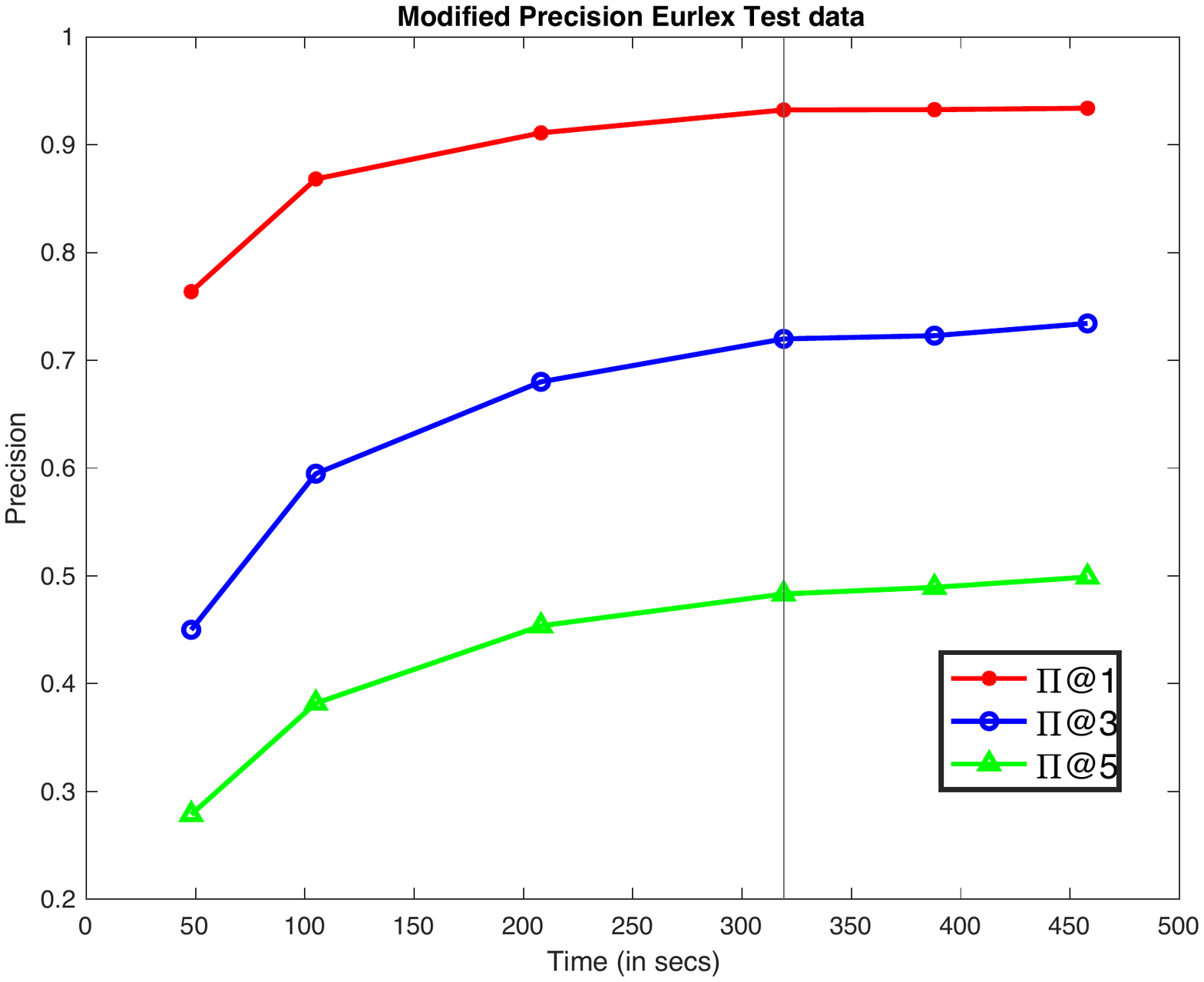}
\includegraphics[height=0.23\textwidth,trim={1cm 7cm 1cm 7cm}]{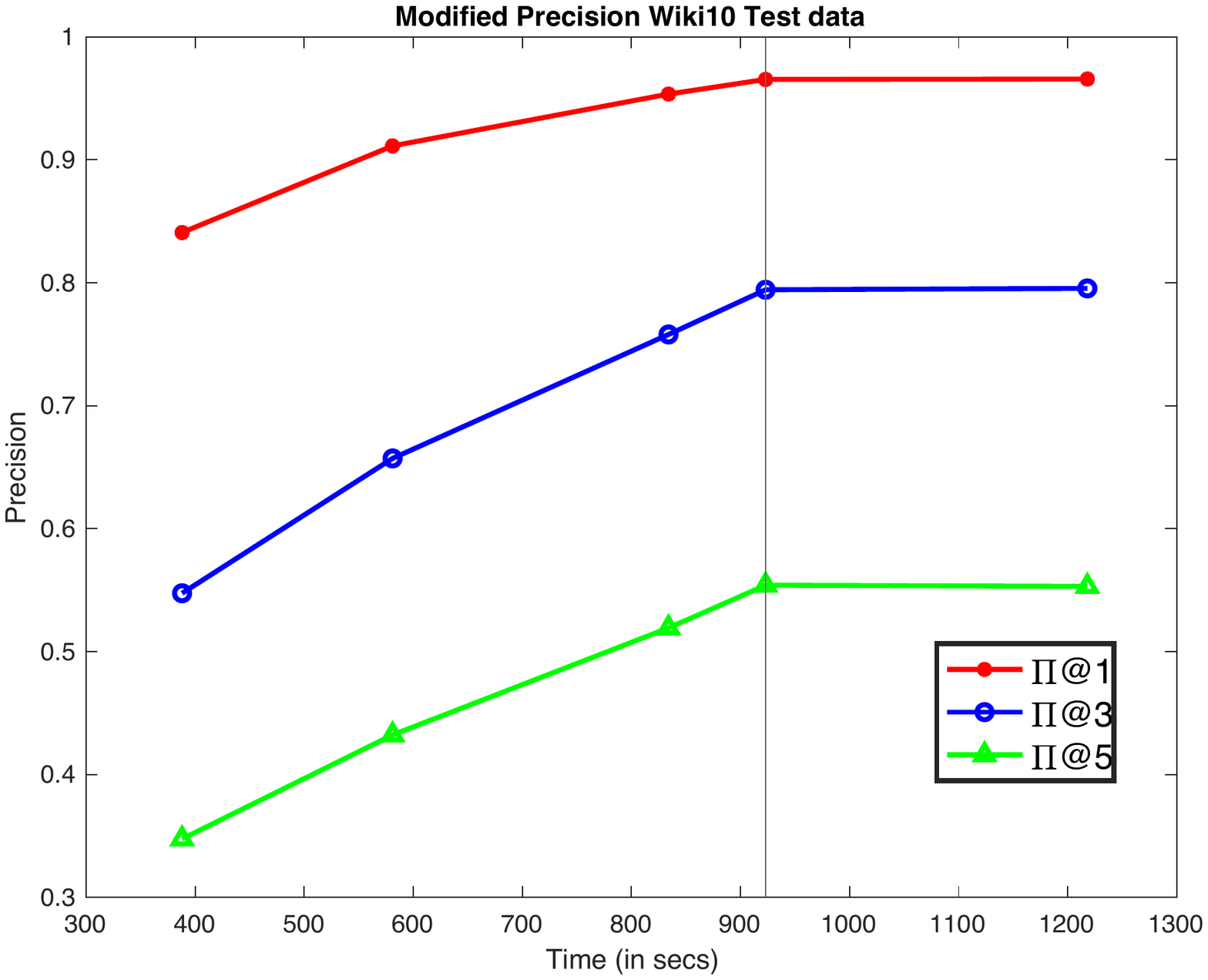}
\includegraphics[height=0.23\textwidth,trim={1cm 7cm 1cm 7cm}]{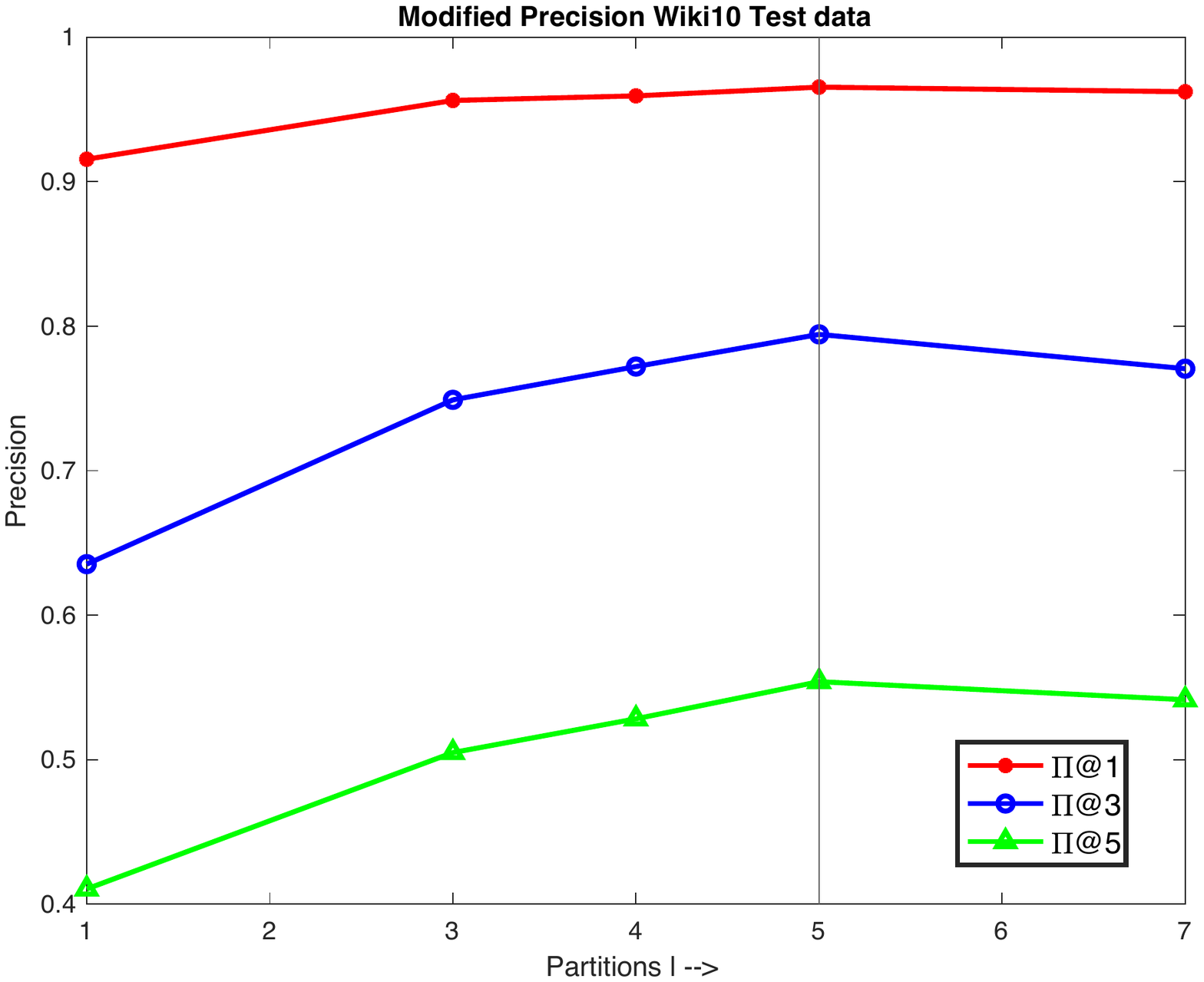}
}
    \caption{Trade off: Precision ($\Pi@1,\Pi@3,\Pi@5$) as  functions of runtimes (in secs) for Eurlex (left) and wiki10 (middle) dataset, and no. of partitions $\ell$ (right).\label{fig:tradeoff}}
    \label{fig:my_label}
\end{figure}

\paragraph{Trade off and improvements:}  In the above two tables, the precision and runtime results we present are based on an optimal trade-off (between accuracy and runtimes) we obtained by our method. In particular, the results reported in the tables are for the smallest $m$ for which our accuracy is close to the SOA tree methods. In figure~\ref{fig:tradeoff}, we plot the precision values ($\Pi@1,\Pi@3,\Pi@5$) as a function of the runtimes (in secs) for Eurlex (left) and wiki10 (middle) datasets, by increasing \# groups $m$ in each partition. Indeed, we notice a clear trade-off: as we increase runtimes, accuracy improves. But beyond a point, the accuracy gain is limited as $m$ is increased. Improved accuracy can be achieved for  higher runtimes (when $m$ is much more than $k\log d$). We also plot $\Pi@k$ versus \# partitions $\ell$ for Wiki10 (right). For smaller $\ell$, it is hard to compute a good NMF for large matrices, and with many partitions, we will miss certain label correlations.
In the tables above, we report the precision numbers -- marked by the black vertical lines in the plots -- that yield the best trade-offs.

  \begin{figure}[tb!]
 \centering
\includegraphics[height=0.3\textwidth,trim={1cm 1cm 1cm 0cm}]{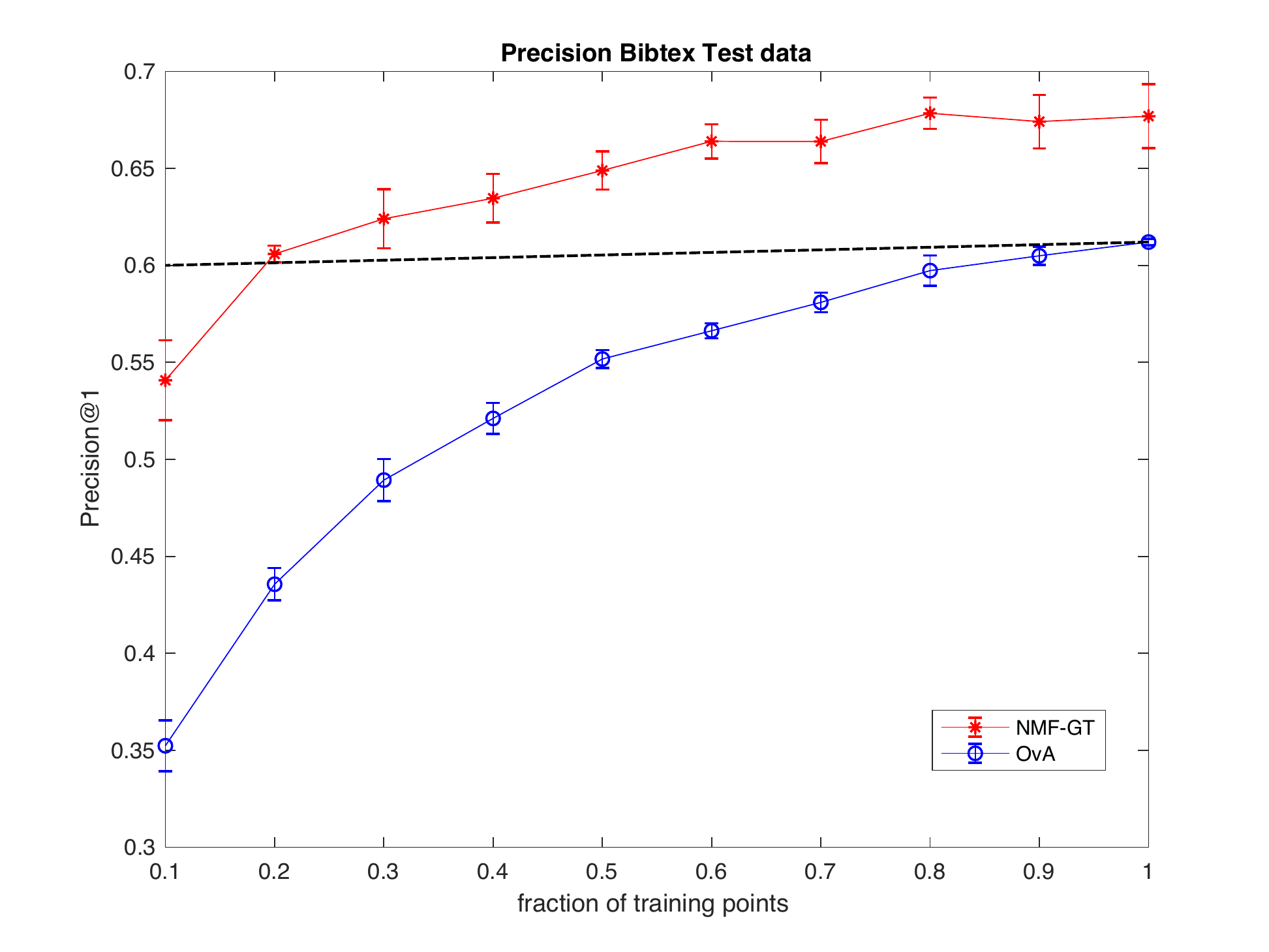}
\includegraphics[height=0.3\textwidth,trim={1cm 1cm 1cm 0cm}]{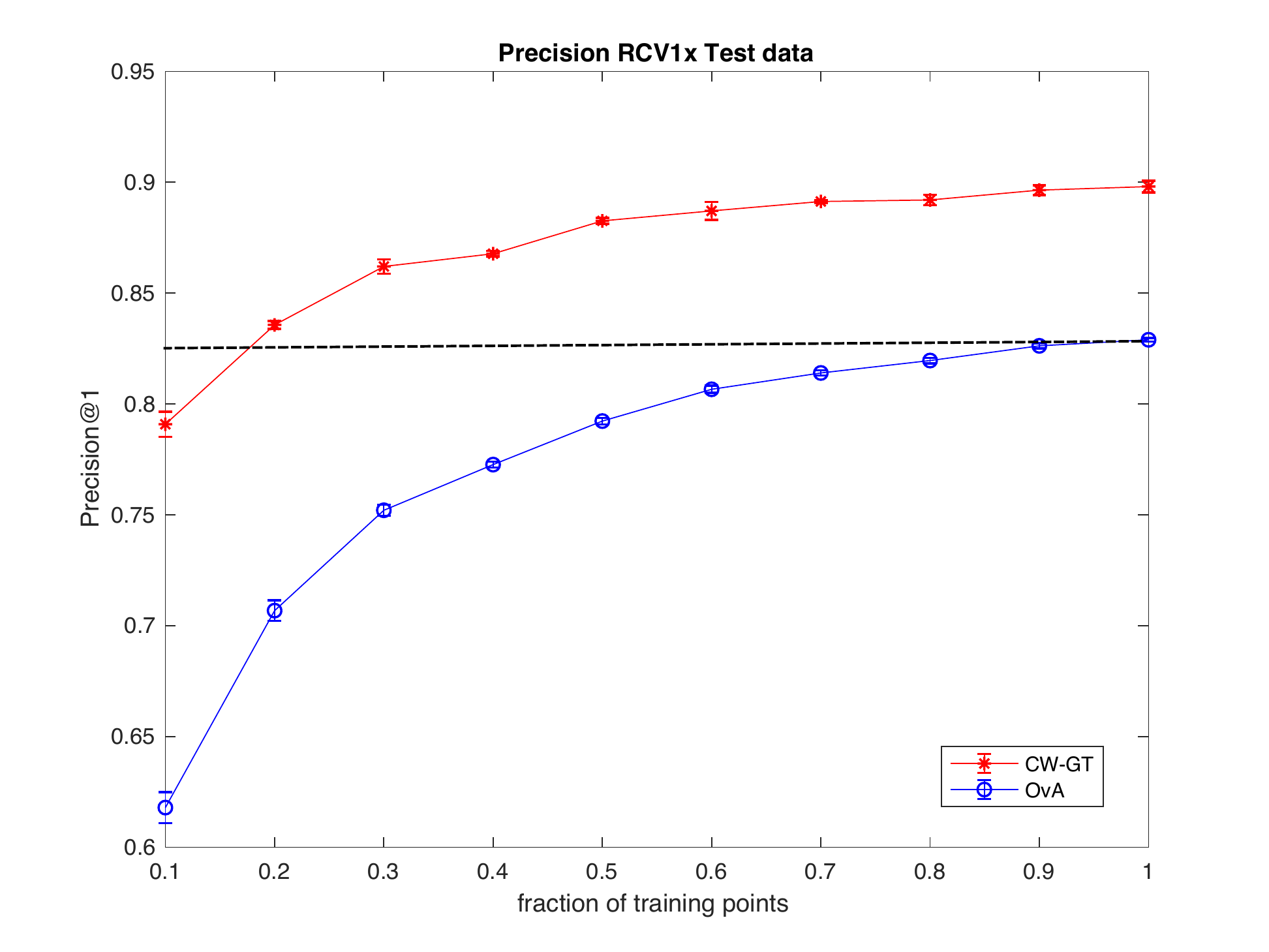}

\caption{Prec@1  for  test data instances for bibtex (left) and RCV1x (right) datasets as a function of fraction of training data used. Error bar over 5 trials} 
\label{fig:2}
\vskip -0.1in
\end{figure}
\paragraph{Learning with less training data:}
 In supervised learning problems such as MLC, training highly accurate models
requires large volumes of labeled data, and creating such volumes of labeled data can be very expensive in many applications~\cite{klein2012101,xu2016robust}. As a result,  there is an increasing interest among research agencies in developing learning algorithms that achieve `Learning with Less Labels' (LwLL)\footnote{\url{darpa.mil/program/learning-with-less-labels}}. Since MLGT requires training  only  $O(k\log d)$ classifiers (as opposed to $d$ classifiers in OvA or other methods), we will need less labeled data
for training the model. In section~\ref{sec:expts}, we present preliminary results that demonstrate how MLGT achieves \emph{learning with less data} for MLC.

Here, we present preliminary results that demonstrate 
how MLGT achieves more accurate (higher precision) with less training data compared to the OvA method (see Table~\ref{tab:table1} in suppl).
Figure~\ref{fig:2} plots the precision (Prec@1) for  test data instances for the bibtex (left) and RCV1x (right) datasets, when different fractions of training data were used to train the MLGT and OvA models. We note that MLGT achieves the same accuracy as OvA with only 15-20\% of the number of training points (over $5\times$ less training data). We used the same binary classifiers for both methods, and MLGT requires only $O(k \log d)$ classifiers, as opposed to OvA, which needs $d$ classifiers. Therefore, MLGT likely requires fewer training data instances.

\subsection*{Conclusions}
In this paper, we extended the MLGT framework~\cite{ubaru2017multilabel} and presented new GT constructions (constant weight and data dependent), and a fast prediction algorithm that requires logarithmic time in the number of labels $d$. We then presented a hierarchical partitioning approach to scale the MLGT approach to larger datsets.  Our computational results show that the NMF construction yields superior performance compared to other 
GT matrices. We also presented a theoretical analysis which showed why the proposed data dependent method (with a non-trivial data-dependent sampling approach) will perform well. With a comprehensive set of experiments, we showed that our method is significantly faster in both training and test times, and yet yields competitive  results compared to other popular XML methods. 

\section*{Broader Impact}
Our work presents a novel approach to solve the extreme multilabel (XML) classification problem. The main contributions of our paper are (a) the use of matrix reordering techniques to hierarchically partition the label space, and (b) the development of a data-dependent group testing scheme, that improves label grouping significantly for MLGT, and can leverage the recently proposed log-time decoding algorithm. These innovations lead to a significantly faster training algorithm compared to most existing methods that yields comparable results. 
The XML problem is encountered in a number of applications from related searches to ad recommendations to natural language understanding in the technology industry, and from gene and molecule classifications to learning neural activities in scientific data analysis. A fast algorithm like ours will enable prediction in high-throughput and real-time settings, and  address some of the limitations in traditional inline search suggestions or approaches for related searches. The preliminary results presented in the supplement demonstrate how the group testing approach achieves \emph{learning with less data} 
for MLC. There is increasing interest among research (and defense) agencies in developing learning algorithms that achieve `Learning with Less Labels' (LwLL), see \url{darpa.mil/program/learning-with-less-labels}.
Thus, the business and research impacts are likely to be significant (and clear).

\paragraph{Acknowledgment:}
We thank the anonymous reviewers for their valuable comments and suggestions.

\begin{thebibliography}{10}

\bibitem{agrawal2013multi}
R.~Agrawal, A.~Gupta, Y.~Prabhu, and M.~Varma.
\newblock Multi-label learning with millions of labels: Recommending advertiser
  bid phrases for web pages.
\newblock In {\em Proceedings of the 22nd international conference on World
  Wide Web}, pages 13--24. ACM, 2013.

\bibitem{babbar2017dismec}
R.~Babbar and B.~Sch{\"o}lkopf.
\newblock Dismec: Distributed sparse machines for extreme multi-label
  classification.
\newblock In {\em Proceedings of the Tenth ACM International Conference on Web
  Search and Data Mining}, pages 721--729. ACM, 2017.

\bibitem{babbar2019data}
R.~Babbar and B.~Sch{\"o}lkopf.
\newblock Data scarcity, robustness and extreme multi-label classification.
\newblock {\em Machine Learning}, 108(8-9):1329--1351, 2019.

\bibitem{Indyk2008}
R.~Berinde, A.~Gilbert, P.~Indyk, H.~Karloff, and M.~Strauss.
\newblock Combining geometry and combinatorics: A unified approach to sparse
  signal recovery.
\newblock {\em CoRR}, abs/0804.4666, 01 2008.

\bibitem{bhatia2015sparse}
K.~Bhatia, H.~Jain, P.~Kar, M.~Varma, and P.~Jain.
\newblock Sparse local embeddings for extreme multi-label classification.
\newblock In {\em Advances in Neural Information Processing Systems}, pages
  730--738, 2015.

\bibitem{bi2013efficient}
W.~Bi and J.~T.~Y. Kwok.
\newblock Efficient multi-label classification with many labels.
\newblock In {\em 30th International Conference on Machine Learning, ICML
  2013}, pages 405--413, 2013.

\bibitem{chang2019modular}
W.-C. Chang, H.-F. Yu, K.~Zhong, Y.~Yang, and I.~Dhillon.
\newblock A modular deep learning approach for extreme multi-label text
  classification.
\newblock {\em arXiv preprint arXiv:1905.02331}, 2019.

\bibitem{chen2012feature}
Y.-N. Chen and H.-T. Lin.
\newblock Feature-aware label space dimension reduction for multi-label
  classification.
\newblock In {\em Advances in Neural Information Processing Systems}, pages
  1529--1537, 2012.

\bibitem{Deng2011}
J.~Deng, S.~Satheesh, A.~C. Berg, and F.~Li.
\newblock Fast and balanced: Efficient label tree learning for large scale
  object recognition.
\newblock In J.~Shawe-Taylor, R.~S. Zemel, P.~L. Bartlett, F.~Pereira, and
  K.~Q. Weinberger, editors, {\em Advances in Neural Information Processing
  Systems 24}, pages 567--575. Curran Associates, Inc., 2011.

\bibitem{gallager62}
R.~{Gallager}.
\newblock Low-density parity-check codes.
\newblock {\em IRE Transactions on Information Theory}, 8(1):21--28, January
  1962.

\bibitem{gupta1997fast}
A.~Gupta.
\newblock Fast and effective algorithms for graph partitioning and
  sparse-matrix ordering.
\newblock {\em IBM Journal of Research and Development}, 41(1.2):171--183,
  1997.

\bibitem{gupta2019distributional}
V.~Gupta, R.~Wadbude, N.~Natarajan, H.~Karnick, P.~Jain, and P.~Rai.
\newblock Distributional semantics meets multi-label learning.
\newblock In {\em Proceedings of the AAAI Conference on Artificial
  Intelligence}, volume~33, pages 3747--3754, 2019.

\bibitem{dingNMF}
C.~H.~Q.~Ding and X.~He.
\newblock On the equivalence of nonnegative matrix factorization and spectral
  clustering.
\newblock In {\em Proceedings of the SIAM International Conference on Data
  Mining}, pages 606--610, 01 2005.

\bibitem{hsu2009multi}
D.~Hsu, S.~M. Kakade, J.~Langford, and T.~Zhang.
\newblock Multi-label prediction via compressed sensing.
\newblock {\em NIPS}, 22:772--780, 2009.

\bibitem{Jain2019slice}
H.~Jain, V.~Balasubramanian, B.~Chunduri, and M.~Varma.
\newblock Slice: Scalable linear extreme classifiers trained on 100 million
  labels for related searches.
\newblock In {\em Proceedings of the Twelfth ACM International Conference on
  Web Search and Data Mining}, WSDM '19, pages 528--536. ACM, 2019.

\bibitem{jain2016extreme}
H.~Jain, Y.~Prabhu, and M.~Varma.
\newblock Extreme multi-label loss functions for recommendation, tagging,
  ranking \& other missing label applications.
\newblock In {\em Proceedings of the 22nd ACM SIGKDD International Conference
  on Knowledge Discovery and Data Mining}, pages 935--944. ACM, 2016.

\bibitem{ijcai2019-361}
A.~Jalan and P.~Kar.
\newblock Accelerating extreme classification via adaptive feature
  agglomeration.
\newblock In {\em Proceedings of the Twenty-Eighth International Joint
  Conference on Artificial Intelligence, {IJCAI-19}}, pages 2600--2606.
  International Joint Conferences on Artificial Intelligence Organization, 7
  2019.

\bibitem{pmlr-v48-jasinska16}
K.~Jasinska, K.~Dembczynski, R.~Busa-Fekete, K.~Pfannschmidt, T.~Klerx, and
  E.~Hullermeier.
\newblock Extreme f-measure maximization using sparse probability estimates.
\newblock In M.~F. Balcan and K.~Q. Weinberger, editors, {\em Proceedings of
  The 33rd International Conference on Machine Learning}, volume~48 of {\em
  Proceedings of Machine Learning Research}, pages 1435--1444, New York, New
  York, USA, 20--22 Jun 2016.

\bibitem{karypis1998fast}
G.~Karypis and V.~Kumar.
\newblock A fast and high quality multilevel scheme for partitioning irregular
  graphs.
\newblock {\em SIAM Journal on scientific Computing}, 20(1):359--392, 1998.

\bibitem{khandagale2019bonsai}
S.~Khandagale, H.~Xiao, and R.~Babbar.
\newblock Bonsai-diverse and shallow trees for extreme multi-label
  classification.
\newblock {\em arXiv preprint arXiv:1904.08249}, 2019.

\bibitem{klein2012101}
A.~Klein and J.~Tourville.
\newblock 101 labeled brain images and a consistent human cortical labeling
  protocol.
\newblock {\em Frontiers in neuroscience}, 6:171, 2012.

\bibitem{symmNMF}
D.~Kuang, C.~Ding, and H.~Park.
\newblock Symmetric nonnegative matrix factorization for graph clustering.
\newblock In {\em Proceedings of the 2012 SIAM International Conference on Data
  Mining}, pages 106--117, 2012.

\bibitem{Saffron2016}
K.~{Lee}, R.~{Pedarsani}, and K.~{Ramchandran}.
\newblock Saffron: A fast, efficient, and robust framework for group testing
  based on sparse-graph codes.
\newblock In {\em 2016 IEEE International Symposium on Information Theory
  (ISIT)}, pages 2873--2877, July 2016.

\bibitem{Liu2017XMLCNN}
J.~Liu, W.-C. Chang, Y.~Wu, and Y.~Yang.
\newblock Deep learning for extreme multi-label text classification.
\newblock In {\em Proceedings of the 40th International ACM SIGIR Conference on
  Research and Development in Information Retrieval}, SIGIR '17, pages
  115--124, New York, NY, USA, 2017.

\bibitem{mcdiarmid1989method}
C.~McDiarmid.
\newblock On the method of bounded differences.
\newblock {\em Surveys in combinatorics}, 141(1):148--188, 1989.

\bibitem{mikolov2013distributed}
T.~Mikolov, I.~Sutskever, K.~Chen, G.~S. Corrado, and J.~Dean.
\newblock Distributed representations of words and phrases and their
  compositionality.
\newblock In {\em Advances in neural information processing systems}, pages
  3111--3119, 2013.

\bibitem{Prabhu2018swift}
Y.~Prabhu, A.~Kag, S.~Gopinath, K.~Dahiya, S.~Harsola, R.~Agrawal, and
  M.~Varma.
\newblock Extreme multi-label learning with label features for warm-start
  tagging, ranking \&\#38; recommendation.
\newblock In {\em Proceedings of the Eleventh ACM International Conference on
  Web Search and Data Mining}, WSDM '18, pages 441--449. ACM, 2018.

\bibitem{Prabhu2018parabel}
Y.~Prabhu, A.~Kag, S.~Harsola, R.~Agrawal, and M.~Varma.
\newblock Parabel: Partitioned label trees for extreme classification with
  application to dynamic search advertising.
\newblock In {\em Proceedings of the 2018 World Wide Web Conference}, WWW '18,
  pages 993--1002, Republic and Canton of Geneva, Switzerland, 2018.
  
\bibitem{Prabhu20141fastxml}
Y.~Prabhu and M.~Varma.
\newblock Fastxml: A fast, accurate and stable tree-classifier for extreme
  multi-label learning.
\newblock In {\em Proceedings of the 20th ACM SIGKDD International Conference
  on Knowledge Discovery and Data Mining}, KDD '14, pages 263--272. ACM, 2014.

\bibitem{saad2003iterative}
Y.~Saad.
\newblock {\em Iterative methods for sparse linear systems}, volume~82.
\newblock siam, 2003.

\bibitem{siblini2018craftml}
W.~Siblini, P.~Kuntz, and F.~Meyer.
\newblock Craftml, an efficient clustering-based random forest for extreme
  multi-label learning.
\newblock In {\em International Conference on Machine Learning}, pages
  4664--4673, 2018.

\bibitem{tai2012multilabel}
F.~Tai and H.-T. Lin.
\newblock Multilabel classification with principal label space transformation.
\newblock {\em Neural Computation}, 24(9):2508--2542, 2012.

\bibitem{trohidis2008multi}
K.~Trohidis.
\newblock Multi-label classification of music into emotions.
\newblock In {\em 9th International Con- ference on Music Information
  Retrieval}, pages 325–-- 330, 2008.

\bibitem{tsoumakas2008effective}
G.~Tsoumakas, I.~Katakis, and I.~Vlahavas.
\newblock Effective and efficient multilabel classification in domains with
  large number of labels.
\newblock In {\em ECML/PKDD 2008 Workshop on Mining Multidimensional Data
  (MMD’08)}, 2008.

\bibitem{ubaru2017multilabel}
S.~Ubaru and A.~Mazumdar.
\newblock Multilabel classification with group testing and codes.
\newblock In {\em Proceedings of the 34th International Conference on Machine
  Learning-Volume 70}, pages 3492--3501. JMLR. org, 2017.

\bibitem{ubaru2016group}
S.~Ubaru, A.~Mazumdar, and A.~Barg.
\newblock Group testing schemes from low-weight codewords of {BCH} codes.
\newblock In {\em Information Theory (ISIT), 2016 IEEE International Symposium
  on}, pages 2863--2867. IEEE, 2016.

\bibitem{symcord}
A.~{Vandaele}, N.~{Gillis}, Q.~{Lei}, K.~{Zhong}, and I.~{Dhillon}.
\newblock Efficient and non-convex coordinate descent for symmetric nonnegative
  matrix factorization.
\newblock {\em IEEE Transactions on Signal Processing}, 64(21):5571--5584,
  2016.

\bibitem{fast2017}
A.~Vem, N.~Thenkarai~Janakiraman, and K.~Narayanan.
\newblock Group testing using left-and-right-regular sparse-graph codes.
\newblock {\em arxiv.org:1701.07477.}, 2017.

\bibitem{wang2009multi}
C.~Wang, S.~Yan, L.~Zhang, and H.-J. Zhang.
\newblock Multi-label sparse coding for automatic image annotation.
\newblock In {\em Computer Vision and Pattern Recognition, 2009. CVPR 2009.
  IEEE Conference on}, pages 1643--1650. IEEE, 2009.

\bibitem{wydmuch2018no}
M.~Wydmuch, K.~Jasinska, M.~Kuznetsov, R.~Busa-Fekete, and K.~Dembczynski.
\newblock A no-regret generalization of hierarchical softmax to extreme
  multi-label classification.
\newblock In {\em Advances in Neural Information Processing Systems}, pages
  6355--6366, 2018.

\bibitem{xu2016robust}
C.~Xu, D.~Tao, and C.~Xu.
\newblock Robust extreme multi-label learning.
\newblock In {\em Proceedings of the 22nd ACM SIGKDD International Conference
  on Knowledge Discovery and Data Mining}, pages 1275--1284. ACM, 2016.

\bibitem{yen2017ppd}
I.~E.-H. Yen, X.~Huang, W.~Dai, P.~Ravikumar, I.~Dhillon, and E.~Xing.
\newblock Ppdsparse: A parallel primal-dual sparse method for extreme
  classification.
\newblock In {\em Proceedings of the 23rd ACM SIGKDD International Conference
  on Knowledge Discovery and Data Mining}, KDD, pages 545--553, 2017.

\bibitem{Yen2016pdsparse}
I.~E.~H. Yen, X.~Huang, K.~Zhong, P.~Ravikumar, and I.~S. Dhillon.
\newblock Pd-sparse: A primal and dual sparse approach to extreme multiclass
  and multilabel classification.
\newblock In {\em Proceedings of the 33rd International Conference on
  International Conference on Machine Learning - Volume 48}, ICML'16, pages
  3069--3077. JMLR.org, 2016.

\bibitem{you2018attentionxml}
R.~You, S.~Dai, Z.~Zhang, H.~Mamitsuka, and S.~Zhu.
\newblock Attentionxml: Extreme multi-label text classification with
  multi-label attention based recurrent neural networks.
\newblock {\em arXiv preprint arXiv:1811.01727}, 2018.

\bibitem{yu2014large}
H.-f. Yu, P.~Jain, P.~Kar, and I.~Dhillon.
\newblock Large-scale multi-label learning with missing labels.
\newblock In {\em Proceedings of the 31st International Conference on Machine
  Learning (ICML-14)}, pages 593--601, 2014.

\bibitem{Zhang2018}
M.-L. Zhang, Y.-K. Li, X.-Y. Liu, and X.~Geng.
\newblock Binary relevance for multi-label learning: an overview.
\newblock {\em Frontiers of Computer Science}, 2018.

\bibitem{Zhang2018Deep}
W.~Zhang, J.~Yan, X.~Wang, and H.~Zha.
\newblock Deep extreme multi-label learning.
\newblock In {\em Proceedings of the 2018 ACM on International Conference on
  Multimedia Retrieval}, ICMR '18, pages 100--107, New York, NY, USA, 2018.
  ACM.

\bibitem{zhang2011multi}
Y.~Zhang and J.~G. Schneider.
\newblock Multi-label output codes using canonical correlation analysis.
\newblock In {\em AISTATS}, pages 873--–882, 2011.

\end{thebibliography}

\newpage
\appendix
\begin{center}
   \bf \large{Supplementary material - Multilabel Classification by Hierarchical Partitioning and Data-dependent
Grouping}
\end{center}
\vskip 0.3in

\section{Constant Weight Construction}
In this supplement, we first describe two constant weight constructions, where each group has the same number of labels, and each label belongs to the same number of groups. 
Such constructions have been shown to perform well in the group testing problem~\cite{ubaru2016group,fast2017}. 
\subsection{Randomized construction}\label{sec:random}

The first construction we consider is based on LDPC (low density parity) codes.  
Gallagher proposed a low density code with constant weights  in~\cite{gallager62}. We can develop a constant weight GT matrix $A$ based on this LPDC construction as follows: Suppose the matrix $A$ we desire has $d$ columns with constant  $c$ ones in each column, and $r$ ones in each row. The LDPC matrix will have $dc/r $ rows in total. The matrix is divided into $c$ submatrices, each containing a single $1$ in each column. The first of these submatrices contains all the ones in descending order, i.e., the $i$th row will have ones in the columns $(i-1)r+1$ to $ir$. The remaining submatrices are simply column permutations of the first. We consider this construction in our experiments.

\subsection{SAFFRON construction}
Recently, in~\cite{Saffron2016},  a biparitite graph based GT construction called SAFFRON (Sparse-grAph  codes Framework For gROup testiNg) was proposed. 
\cite{fast2017} extended this SAFFRON construction to form  left-and-right-regular sparse-graph codes called regular-SAFFRON. The adjacency matrices corresponding to such graphs give us the desired  constant weight constructions. 
The regular-SAFFRON construction starts with a  left-and-right-regular  graph $G_{cr}(d,m_1)$, with $d$ left nodes called variable nodes,  and $m_1$ right nodes called bin nodes. The $d.c$ edge  connections from  the  left and $m_1.r$ edge connections from the right are paired up according to a random permutation.

Let $T_G\in \{0,1\}^{m_1\times d}$ be the adjacency matrix corresponding to the  left-and-right-regular  graph $G_{cr}(d,m_1)$. Then, $T_G$ has $c$ ones in each column and $r$ ones in each row. Let $U \in \{0,1\}^{m_2\times d}$ be the universal signature matrix (see~\cite{Indyk2008,fast2017} for definition). If $t_i$ is the $i$the row of $T_G = [t_1^T, \ldots , t_{m_1}^T]^T$, then the GT matrix A is formed as $A = [A_1^T, \dots, A_{m_1}^T]^T$, where the submatrix 
$A_i = Udiag(t_i)$ of size $m_2\times d$. The total tests will be $m = m_1\cdot m_2$. We have the following recovery guarantee of this construction:

   \begin{proposition}\label{prop:1}
   Suppose we wish to recover a $k$ sparse binary vector $y\in\{0,1\}^d$.
   A  binary testing matrix $A$  formed from the regular-SAFFRON graph with 
    $m = \tau_1.k \log \frac{d}{k}$ tests
   recovers $1-\veps$ proportion of the support of  $y$ correctly
    with high probability (w.h.p), for any $\veps>0$. With $m = \tau_2 k \log k  \log\frac{d}{k}$, we can recover the whole support
    set w.h.p. The constants $\tau_1$ and $\tau_2$ depend on $c,r$ and the error tolerance $\veps$.
     The 
   computational complexity of the  decoding scheme will be $O(k \log \frac{d}{k})$.
   \end{proposition}
Proof of the proposition can be found in~\cite{fast2017}. The decoding algorithm was discussed in the main text.

\section{Proof of Theorem~\ref{theo:1}}

Next, we sketch the proof of Theorem~\ref{theo:1} in the main text.
\begin{proof}

Let us denote the entries of $\tilde{H}$ and $A$ as $\tilde{h}_{i,j}$ and $a_{i,j}$ respectively, $i = 1, \dots, m; j = 1, \dots, n$. From our construction: $\Pr(a_{i,j} =1) = \tilde{h}_{i,j}$ and $\Pr(a_{i,j} =0) = 1-\tilde{h}_{i,j}.$

First, let us find the probability that $z_i=0.$ Since $z_i$ will be 0 if and only if the support of $i$th row of $A$ has no intersection with the support of $y$, hence,
$$
\Pr(z_i =0) = \prod_{j \in \supp(y)} \Pr(a_{i,j} =0) =  \prod_{j \in \supp(y)} (1-\tilde{h}_{i,j}). 
$$
Now note that, $b_j = \sum_{i=1}^m a_{i,j} z_i$. 
Therefore, $\avg[{b_j}] = \sum_{i=1}^m \avg[a_{i,j} z_i] =  \sum_{i=1}^m\Pr(a_{i,j} z_i=1). $
It turns out that,
\begin{align*}
\Pr(a_{i,j} z_i=1) &= \Pr(a_{i,j} =1, z_i=1) \\
&= \Pr(a_{i,j} =1) \Pr(z_i=1 \mid a_{i,j} =1)\\
& = \tilde{h}_{i,j}(1 -  \Pr(z_i=0 \mid a_{i,j} =1)).
\end{align*}
Now, we consider two cases. When $j \in \supp(y)$, $\Pr(z_i=0 \mid a_{i,j} =1) = \Pr(\forall j \in \supp(y), a_{i,j} =0 \mid a_{i,j} =1) = 0$. On the other hand, when $j \notin \supp(y)$, $\Pr(z_i=0 \mid a_{i,j} =1) = \Pr(z_i=0) = \prod_{l \in \supp(y)} (1-\tilde{h}_{i,l})$. Therefore,
\[
\Pr(a_{i,j} z_i=1) = \begin{cases}
\tilde{h}_{i,j}, & j \in \supp(y)\\
 \prod_{l \in \supp(y)} (1-\tilde{h}_{i,l}) & j \notin \supp(y).
 \end{cases}
\]
Hence, when $j \in \supp(y)$,
$$
\avg[b_j] = \sum_{i=1}^m \Pr(a_{i,j}z_i =1) = \sum_{i=1}^m \tilde{h}_{i,j} =c.
$$

But when $j \notin \supp(y)$,
\begin{small}
\begin{align*}
\avg[b_j] &=\sum_{i=1}^m \Pr(a_{i,j}z_i =1) \\
& = \sum_{i=1}^m  \prod_{l \in \supp(y)} (1-\tilde{h}_{i,l}) \le  \sum_{i=1}^m  \prod_{l \in \supp(y)} \exp(-\tilde{h}_{i,l})\\
& =  \sum_{i=1}^m \exp\Big(-\sum_{l \in \supp(y)}\tilde{h}_{i,l}\Big) = \sum_{i=1}^m \exp(-\langle y, \tilde{h}^{(i)}\rangle).
\end{align*}
\vskip-.75cm
\end{small}
\end{proof}
We can make stronger claims to bolster this theorem. Since the random variables $b_j, j =1, \dots, n$ are all Lipschitz functions of independent underlying variables, by using McDiarmid inequality~\cite{mcdiarmid1989method} we can say that they are tightly concentrated around their respective average values.

\section{Additional experimental results}
Here, we  present additional results and further discuss the results we presented in the main text for the proposed methods.
We then give few results which help us  better understand the parameters that affect the performance of our MLGT method. First, we describe the evaluation metrics used in the main text and here for comparison.

\paragraph{Results discussion:} 
In table~\ref{tab:table2} of main text, we summarized the results obtained for six methods for different datasets.  
We note that NMF-GT performs very  well given its low computational burden. 
PfastreXML and Parabel, on the other hand, yield slightly more accurate results but require significantly longer run times. 

Note that, when compared to the MLGT, the other  methods  require significantly more time for training. This is because, the tree based methods use k-means clustering recursively to build the label tree/s, and require several OvA classifiers to be trained, one per each label in the leaf nodes. 
OvA methods are obviously expensive since they learn $d$ number of classifiers.
Moreover, the prediction time for MLGT is also orders of magnitude less than many of the popular methods. 
In addition, the other methods have several parameters that need  to be  tuned (we used the default settings provided by the authors). 
We also  note that the main routines of most other methods are written in C/C++ language, while  MLGT was implemented in Matlab and hence the run times can be further improved to enable truly real-time predictions.

In Table~\ref{tab:table3} of the main paper, for the large two datasets, the label set was divided into blocks of sizes roughly around $40K$. We also used negative sampling of the training data for each block as done in many recent XML works~\cite{Prabhu2018parabel,Jain2019slice}. We also reduced the feature dimension via. sketching. For hierarchical partitioning, we used the vertex separator approach described in the main text, using the FORTRAN code provided by the author of~\cite{gupta1997fast}. The reordering for the four datasets in Table 3 are given in Figure~1 for the main text. The approach is extremely fast, and the runtime for the four datasets for reordering and partitioning were: 

\emph{Eurlex: 0.5s; Wiki10: 4.11s; WikiLSHTC: 40.3s; and Amazon670: 15.5s.}

For Eurlex and Wiki10, the accuracy and runtime results for SLEEC, PfastreXML and Parabel were computed by us using their matlab codes. Results for these three methods for the remain two datasets, and all results for the additional four methods (Dismec, PPD-sparse,  XT and XML-CNN) were obtained from~\cite{Prabhu2018parabel} and~\cite{wydmuch2018no}.
All runtimes are based on single core implementation.

\begin{figure*}[tb]
\centering
\includegraphics[height=0.23\textwidth,trim={1cm 1cm 1cm 1cm}]{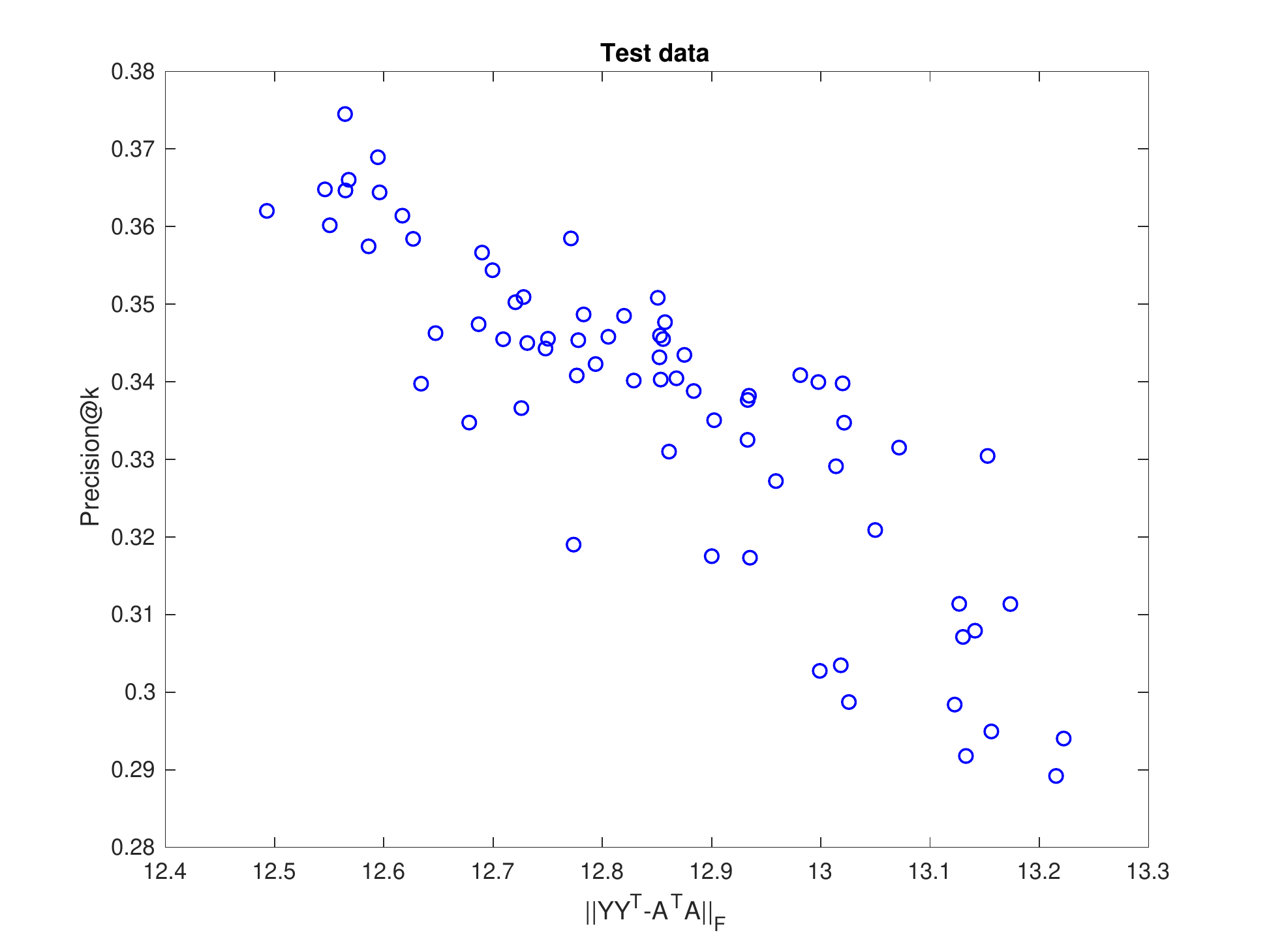}
\includegraphics[height=0.23\textwidth,trim={1cm 1cm 1cm 1cm}]{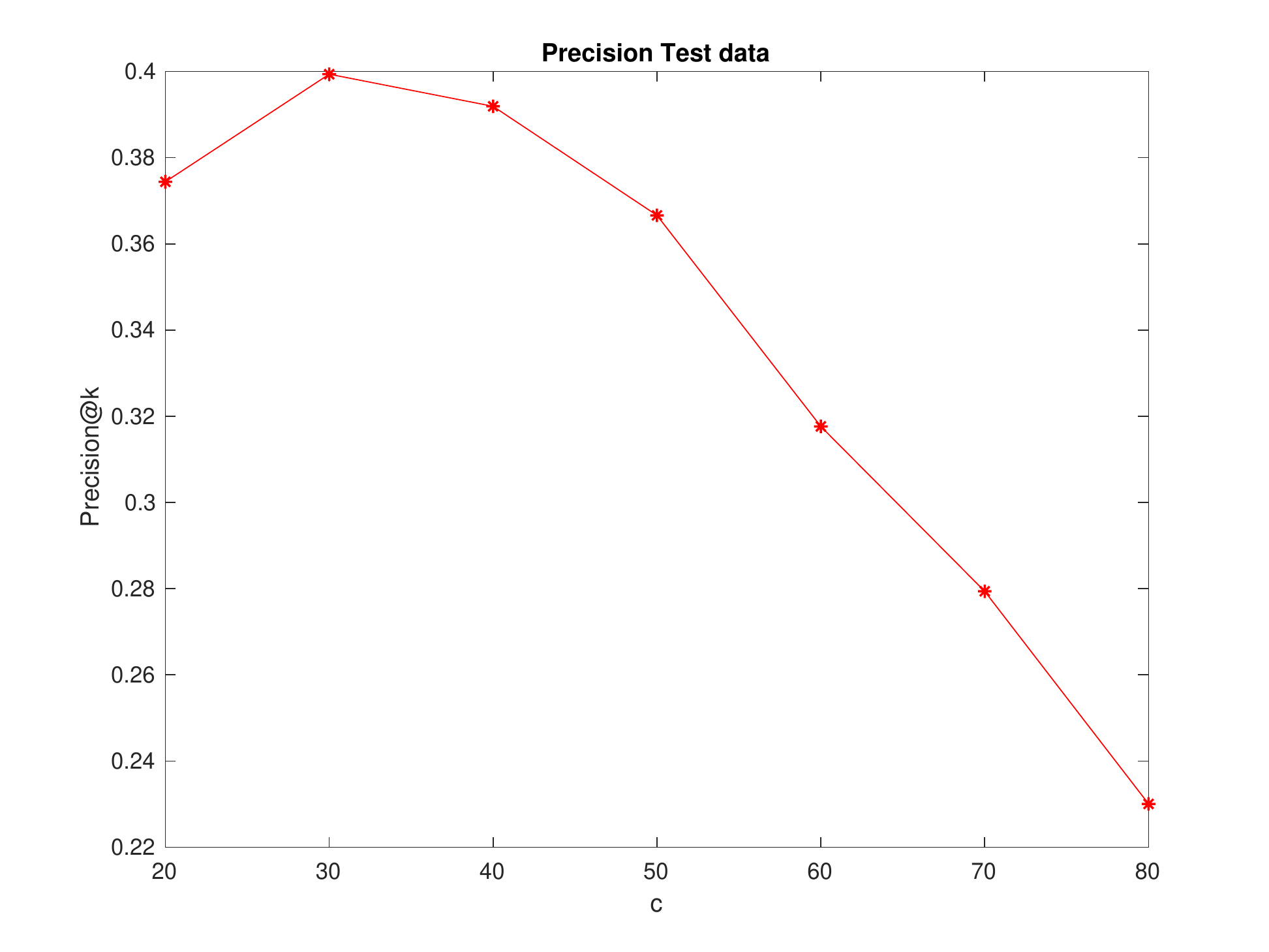}
\includegraphics[height=0.23\textwidth,trim={1cm 1cm 1cm 1cm}]{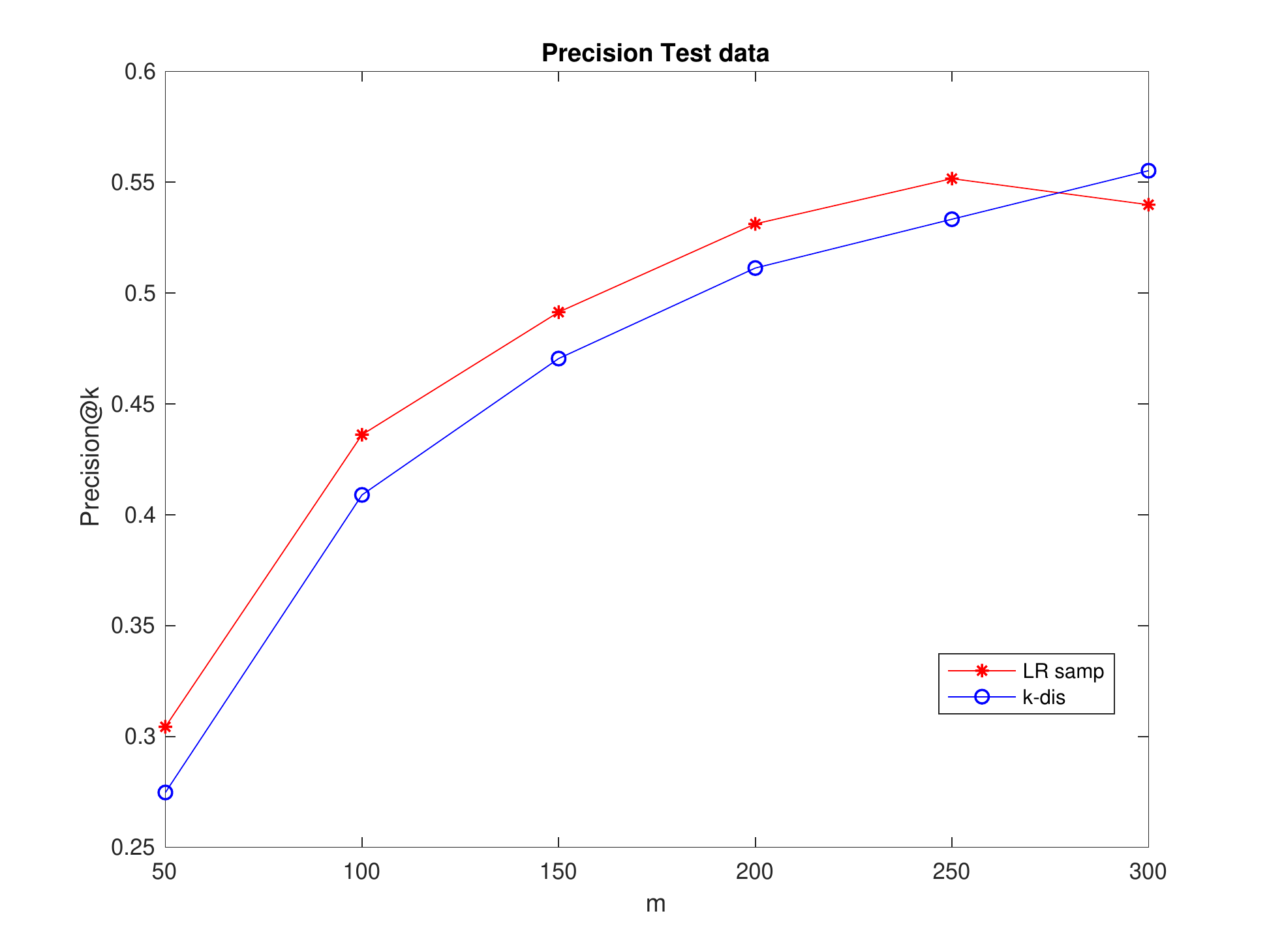}
\caption{Analysis: (Left) Relation between P@k and the correlation metric $\Phi_Y(A)$, (Middle)  Relation between P@k and
column sparsity c, and (Right) Performance of NMF for larger $m$.} 
\label{fig:3}
\end{figure*}

\paragraph{MLGT Analysis:}
We conducted several numerical tests to analysis the performance of MLGT with respect to various settings. Figure~\ref{fig:3} presents few of these numerical analysis results, which help us understand the performance of MLGT better.  In the left figure, we plot the P@k achieve by MLGT with different GT constructions, as a function  of the 
 the correlation metric $\Phi_Y(A)$. The different points (circle) in the plot correspond to different GT matrices with different  $\Phi_Y(A)$. These GT matrices were formed by randomly permuting $k$-disjunct matrices, and changing its size.
 We observe that GT matrices with lower $\Phi_Y(A)$, yield better classification. These results motivated us to develop the data-dependent grouping approach.
 
 In the middle plot, we have the performance of the NMF-GT method for different column sparsity $c$. We clearly note that as $c$ increases, the performance first increases, and then reduces for larger $c$. This is because, for larger $c$, the GT matrix will have higher coherence between the columns. As indicated in our analysis, the performance of the GT   construction will depends on this coherence. This analysis motivated us to use the search technique described in Remark 1, to select the optimal column sparsity~$c$.
 
 In the right plot, we compare the performance of NMF-GT vs CW-GT as a function of number of groups $m$ for the Eurlex dataset. We observe that for smaller $m$, NMF-GT performs better. However, for larger $m$ and more so for larger number of label $d$,  NMF-GT becomes less accurate. This is due to the difficulty in computing accurate NMF for such large matrices.
 NMF is known to be an NP hard problem. This result likely explains why the NMF-GT's performance on larger datasets is less accurate. A possible approach to improve the accuracy of NMF-GT is to use the Hierarchical approach described above
 and split the large label set into smaller disjoint subsets, and apply NMF-GT independently.

    \begin{table}[t]
 \caption{Average Hamming loss errors in reduction v/s training}\label{tab:tab3}
 \begin{center} 
  {\footnotesize
 \begin{tabular}{|l|c|c|c|c|c|}
\hline
Dataset&  &   \multicolumn{2}{c|}{NMF-GT} &   \multicolumn{2}{c|}{CW-GT}\\\cline{3-6}

&$d$&       R-Loss & T-Loss &  R-Loss & T-Loss  \\
\hline
Bibtex 		&159&3.49&3.68&2.95&4.30\\
RCV1-2K	&2016&3.99&4.72&3.96&4.91\\
EurLex-4K 	&3993&1.38&4.77&1.05&5.03\\
\hline
\end{tabular}
}
 \end{center} 
\vskip -0.1in
\end{table}

 In table~\ref{tab:tab3}, we list the average Hamming loss errors we suffer in label reduction (and decoding) 
 when using NMF-GT and CW-GT for the three datasets. That is, we check the average error in the group testing procedure 
 (label reduction and decoding), without classifiers. 
 We also list the average Hamming loss in the training data after classification for comparison. 
 We observe that, the NMF-GT has worse reduction loss compared to CW-GT. This is because, NMF-GT is data dependent, and is not close to being k-disjunct as oppose  to CW-GT, which  is random.
 However, we note that the training loss of NMF-GT is better. This shows that, even though the reduction-decoding is imperfect (introduces more noise), NMF-GT results in better individual classifiers. These comparisons show that data-dependent grouping will indeed result in improved classifiers.
 
 \paragraph{Implementation details:}
 All experiments for NMFGT and He-NMFGT were implemented in Matlab, and conducted on a standard work station with Intel i5 core 2.3GHz machine. 
 The timings reported were computed using the \texttt{cputime} function in Matlab. For the SLEEC method, we could not compute $\Pi@k$ as in eq.~\ref{eq:Pk}, since the source code did not output the score matrix. The $\Pi@k$ reported for SLEEC in Table 4 were the P@k returned by source code. Also, for the last 2 examples, SLEEC was run for 50 iterations (for the rest it was 200).

 \end{document}